\documentclass[twoside]{article}

\usepackage[accepted]{aistats2024}
\usepackage[round]{natbib}

\usepackage{textcomp}

\usepackage{hyperref}       
\usepackage{url}            
\usepackage{booktabs}       
\usepackage{amsfonts}       
\usepackage{nicefrac}       
\usepackage{xcolor}         
\usepackage{microtype}

\usepackage{graphicx}
\usepackage{multirow}

\usepackage{caption}
\usepackage{subcaption}
\usepackage{scrextend}

\usepackage{amsmath}
\usepackage{amsthm}
\usepackage{amssymb}
\usepackage{pifont}

\usepackage{lineno}
\usepackage{algorithm2e}
\usepackage{float}
\RestyleAlgo{ruled}

\newtheorem{theorem}{Theorem}

\newtheorem{remark}{Remark}
\newtheorem{proposition}{Proposition}
\newtheorem{definition}{Definition}

\newtheorem{cor}{Corollary}

\newcommand{\expect}{{\bf E}}
\newcommand{\bw}{{\bf w}}
\newcommand{\bu}{{\bf u}}
\newcommand{\bW}{{\bf W}}
\newcommand{\bv}{{\bf v}}

\newcommand{\bV}{{\bf V}}

\newcommand{\cmark}{\ding{51}}%
\newcommand{\xmark}{\ding{55}}%

\begin{document}

%

%

\twocolumn[

\aistatstitle{Fair Supervised Learning with A Simple Random Sampler of Sensitive Attributes}


\aistatsauthor{ Jinwon Sohn \And Qifan Song \And  Guang Lin }

\aistatsaddress{ Purdue University \And  Purdue University \And Purdue University } ]

\begin{abstract}
  As the data-driven decision process becomes dominating for industrial applications, fairness-aware machine learning arouses great attention in various areas. This work proposes fairness penalties learned by neural networks with a simple random sampler of sensitive attributes for non-discriminatory supervised learning. In contrast to many existing works that critically rely on the discreteness of sensitive attributes and response variables, the proposed penalty is able to handle versatile formats of the sensitive attributes, so it is more extensively applicable in practice than many existing algorithms. This penalty enables us to build a computationally efficient group-level in-processing fairness-aware training framework. Empirical evidence shows that our framework enjoys better utility and fairness measures on popular benchmark data sets than competing methods. We also theoretically characterize estimation errors and loss of utility of the proposed neural-penalized risk minimization problem.
\end{abstract}

\section{INTRODUCTION}
\label{sec:intro}

Algorithmic fairness has been a growing research area as the prediction-based decision process becomes more and more prevalent. The legal examples include the US Equal Credit Opportunity Act, the European Union's General Data Protection Regulation, and the Fair Credit Reporting Act, to name a few. On the academic side, \cite{mhas:etal:21} discussed the importance and challenges of algorithmic fairness in public health. \cite{kozo:etal:22} studied various mitigation strategies for the credit scoring application. 
\cite{dear:etal:22} gave an overview of potential areas demanding fairness-aware business analytics with intriguing real-world examples such as dynamic pricing, distribution of vaccines, job applications, and so forth. Similar motivating examples can be easily found in other fields as well, such as education \citep{louk:etal:19}, finance \citep{das:etal:21}, mortgage lending \citep{lee:flor:21}, and computational medicine \citep{xu:etal:22}.

In pursuit of the rising demand for mitigating societal bias in decision-making processes, there have been versatile approaches, which are mainly categorized into pre-processing, in-processing, and post-processing \citep{baro:etal:17, cato:haas:20,paga:etal:23,xian:etal:23}. Pre-processing includes relabeling, reweighting, or resampling of data instances to ease possible discrimination of a learned model \citep{kami:cald:12}. Synthesizing data to be fair also belongs to this class as well \citep{xu:etal:18, satt:etal:19, breu:etal:21}. In the field of in-processing, optimization with fairness constraints has been one of the main branches \citep{agar:etal:18,komi:etal:18,agar:etal:19,zafa:etal:19,scut:etal:22,jung:etal:23}. Comparably, model-based fairness control has been also extensively studied, which regularizes the degree of discrimination through an auxiliary model \citep{xie:etal:17,beut:etal:17, zhan:etal:18, adel:etal:19,mary:etal:19,lee:etal:22}. Following the model-agnostic spirit, post-processing directly modifies the model's outcome such that the reporting values are non-discriminatory while minimizing the loss of utility \citep{hard:etal:16, plei:etal:17, zeng:etal:22}. Out of such mitigation categories, there have been versatile studies imposing fairness in clustering \citep{wang:etal:23}, feature selection \citep{quin:etal:22}, reinforcement learning \citep{deng:etal:22}, and so on.

There are three key notions for pursuing better algorithmic fairness: {\bf independence}, {\bf separation}, and {\bf sufficiency} \citep{baro:etal:17,cato:haas:20,paga:etal:23}. Throughout this paper, we denote $X$ as covariates, $Y$ as a response, $A$ as sensitive random variables to be protected (e.g., race or gender), and $h$ as a scoring model that makes a data-driven decision. The independence condition requires that $A$ is independent of $h(X)$, i.e., $A\perp h(X)$; statistical parity is an example of independence. Separation and sufficiency, on the other hand, are defined based on the conditional independence structure, i.e., $A\perp h(X)|Y$  and $A\perp Y|h(X)$ respectively. For instance, if two groups in a binary $A$ have the same false-positive and false-negative errors in a binary classification problem, the circumstance achieves the error rate balance \citep{chou:17} or equalized odds \citep{hard:etal:16}, implying separation. For sufficiency, evaluating the calibration of the model across sensitive factors is a representative example  \citep{chou:17,baro:etal:17}.

Throughout this paper, let $P(X)$ denote the distribution law of a random variable $X$. For generic random variables $X_1$ and $X_2$, $P(X_1)=P(X_2)$ means that $X_1$ and $X_2$ have the same distribution, and $P(X_1|X_2=x_2)$ implies the conditional distribution of $X_1$ given $X_2=x_2$. If the two random variables satisfy $P(X_1,X_2)=P(X_1)P(X_2)$, we say $X_1$ and $X_2$ are independent of each other. On the other hand, $p(x)$ denotes the probability mass/density function of $X$, depending on the context.

In this work, we propose a model-based in-processing framework that flexibly captures and controls underlying societal discrimination realized through the output of $h$. Our contributions are as follows:

\begin{itemize}
    \item First, we devise a fairness penalty by a neural network denoted as $D$ that leverages a simple random sampler of sensitive variables for independence to measure the discrepancy between $P(h(X),A)$ and $P(h(X))P(A)$. The penalty can accommodate various scenarios even when $A$ is a mix of continuous and discrete variables, in which many prior works fail as exhibited in Table~\ref{tab:sum}. 
    \item Second, the penalty is further extended to embrace separation while inheriting all the remarkable properties shown in the independence case.
    \item Third, we mathematically quantify the underlying mechanism of the proposed fairness-controlled supervised learning. Specifically, we derive the upper bounds of estimation error and loss of utility with the imposed penalty based on statistical learning theory.
\end{itemize}

\section{RELATED WORKS}
\label{sec:rel}

\paragraph{Density Matching}
Achieving fairness through density matching has been steadily implemented. \cite{quad:shar:17} designed a privileged learning using a variant of a support vector machine on which a fairness constraint via maximum mean discrepancy is imposed. \cite{cho:etal:20} performed matching distributions between different sensitive groups via a differentiable kernel density estimation technique. \cite{li:etal:21} adopted the loss function used in the generative adversarial network \citep{good:etal:14} as an extra penalty to facilitate the scoring distributions $h$ conditional on different $A$ values to be indistinguishable. These density-matching-based regularizations work well but are usually limited to the case when $A$ and $Y$ are discrete or binary. That is because their backbone optimization structures essentially depend on \emph{sub-groups comparison}. As an example of independence for binary $A$, the loss function relies on two subsets of data, corresponding to $A=1$ and $A=0$ respectively, to measure the distributional difference between sub-groups (i.e., $P(h(X)|A=1)$ vs $P(h(X)|A=0)$). Obviously, this technique does not apply to continuous $A$. \cite{roma:etal:20} suggested using an extra generator that produces $A$ conditionally on $Y$ to ensure separation. Although this approach technically bypasses the sub-group comparison, it critically relies on the quality of the conditional generator. If $A$ is complex, e.g., a mix of categorical and continuous variables, finding a good generator that captures the true distribution could be an extremely difficult task \citep[see, e.g., ][]{xule:etal:19,kote:etal:23}. 


\paragraph{HGR penalty and beyond}
To tackle such limitations in the literature, diverse approaches have been proposed. \cite{mary:etal:19} first utilized Hirschfeld-Gebelein-R\'enyi (HGR) maximal correlation which can capture the nonlinear correlation between $A$ and $h(X)$ regardless of the variable type. \cite{grar:etal:19} and \cite{lee:etal:22} approximated the HGR maximal correlation via neural networks, to achieve independence and separation.
These approaches conduct min-max optimization where the auxiliary neural networks approximate the HGR by maximization and the predictive model $h$ is trained to minimize the approximated HGR. 
The auxiliary neural nets in the training process take variable $A$ as the input instead of fitting $A$, hence the multivariate nature of $A$ does not matter anymore. 
Besides, \cite{du:etal:21} recently proposed a way of neutralizing the hidden layer of a neural-net model, which can also handle multivariate $A$; \cite{scut:etal:22} developed a fair (generalized) linear model that places the ridge penalty to adjust the violation of discrimination for any kinds of $A$ and $Y$. 
\begin{table*}[t]
\centering
\caption{Summary of applicability of in-processing methods: ``NN'' denotes whether or not neural network models are supported. ``Ind.'' and ``Sep.'' imply independence and separation. Abbreviations ``D.'', ``C.'', and ``M.'' imply discrete, continuous, and mixed (continuous and discrete) type; Abbreviations ``$A$'' and ``$Y$'' mean sensitive and outcome variables. ``$\triangle$'' means the framework does not explicitly control specific fairness metrics.  ``$\square$'' means the method is technically applicable but requires a non-trivial modification. ``B.'' implies the framework only supports a binary type variable.
}
\begin{tabular}{l|c|cc|ccccc}
\toprule
Methods & NN & Ind. & Sep.  & D. $A$ & C. $A$ & M. $A$ & D. $Y$ & C. $Y$  \\
\toprule
\citet{edwa:stor:15}     &  \cmark  &  \cmark &  \xmark &  B &  \xmark     & \xmark  &   \cmark   &  \xmark        \\
\citet{agar:etal:18}     &  \cmark  &  \cmark  &   \cmark  &  \cmark &  \xmark     & \xmark  &   \cmark   &  \xmark        \\
\citet{adel:etal:19}     &  \cmark  &  \cmark   &   \cmark  &  B & \xmark      &  \xmark   &   \cmark   &  \xmark    \\
\citet{cho:etal:20}      &\cmark& \cmark & \cmark &    \cmark  &  \xmark     &  \xmark   &  \cmark    &   \xmark      \\
\citet{roma:etal:20}     &\cmark& \xmark & \cmark &    \cmark  &  \cmark     &   $\square$    &  \cmark    &   \cmark      \\
\citet{li:etal:21} &\cmark&\cmark & \cmark &     B    &   \xmark    &  \xmark   &   \cmark   &    \xmark   \\
\citet{du:etal:21} &\cmark& $\triangle$  &   $\triangle$  &     \cmark   &   \cmark    &  \cmark   &   \cmark   &    \xmark     \\
\citet{scut:etal:22} &\xmark&\cmark& \cmark &   \cmark   &   \cmark    &  \cmark   &   \cmark   &    \cmark     \\
\citet{lee:etal:22} &\cmark&\cmark &\cmark  &\cmark     &\cmark      & \cmark    &\cmark     &\cmark   \\

Ours  &\cmark&\cmark &\cmark  &\cmark     &\cmark      &\cmark    &\cmark     &\cmark  \\
\bottomrule
\end{tabular}
\label{tab:sum}
\end{table*}
\vspace{-0.1in}

\section{METHODOLOGY}
\subsection{Penalty with Resampled $A$}
\label{sec:SBP}

Focusing on independence, this section describes the main idea which generalizes to any dimension or type of sensitive variables. The methodological novelty originates from the use of a simple random sampler of $A$ to quantify the degree of fairness via a discriminative neural network $D$. Given a sufficient network capacity, the learned $D$ can capture the violation of statistical independence, so it is used as a penalty (or adversarial) network for the risk-minimization problem of $h$ such that the resulting $h$ becomes fair.

To start with, let's formally define $X\in{\cal X}\subset \mathbb{R}^p$ and $A\in {\cal A}\subset \mathbb{R}^l$ as the multivariate non-sensitive and sensitive random variable respectively, and $Y\in {\cal Y}\subset \mathbb{R}$ as the univariate outcome variable. Given a scoring model $h:{\cal X} \rightarrow {\cal S}\subset \mathbb{R}$ and a loss function $L:{\cal Y}\times {\cal S}\rightarrow {\mathbb R}$, we denote the risk function as $R(h) = \expect_{Y,X} [L(Y,h(X))]$ where $\expect_{Y,X}$ means the expectation w.r.t. the joint distribution of ($X$,$Y$). Note $h$ does not explicitly depend on $A$. This structure conceptually carries on fairness via blinding and does not input $A$ in the inference phase \citep{quad:shar:17,zafa:etal:19}. In order to evaluate independence in general settings (i.e., beyond binary $A$ and $Y$), we generalize the existing notion of statistical parity as follows. 
\begin{definition}[Generalized Statistical Parity (GSP)]
\label{def:gsp}
    The scoring model $h$ is called to satisfy the generalized statistical parity if $P(h(X)|A=a) = P(h(X))$ for all $a \in {\cal A}$.
\end{definition}
GSP is more general than the classical statistical parity (SP); if ${\cal A}=\{0,1\}$ for the binary classification problem, GSP implies SP, i.e., $P(h(X) > \tau|A=1){=}P(h(X) > \tau|A=0)$ for any classification threshold $\tau$.  

Now, we define the penalty that promotes the fairness of $h$. Inspired by the noise-contrastive loss \citep{gutm:hyva:12,good:etal:14}, with $D:{\cal S}\times {\cal A}\rightarrow (0,1)$, let's define $R_F(h)=\sup_D~R_F(h;D)$ where 
\begin{align*}
 R_F(h;D) &= \expect_{X,A} [\log D(h(X),A)] \\
           &+ \expect_{X,A'} [\log (1-D(h(X),A'))], 
\end{align*}
and $A'$ shares the same distribution with $A$ but is independent of $(X,A)$. We remark that $A'$ can be easily obtained by applying the simple random sampling $A'$ from $P(A)$. The detailed implementation for this marginal sampling is discussed in Supplementary~\ref{appen:alg_gsp}. The following proposition theoretically validates that $R_F(h)$ captures the discrepancy between $P(h(X)|A)$ and $P(h(X))$.
\begin{proposition}
\label{prop:sp}
Let $p_{A}$, $p_{h(X)}$, $p_{h(X)|A}$, and  $p_{h(X),A}$ be the marginal densities of $A$ and $h(X)$, the conditional density of $h(X)$ given $A$ respectively, and the joint density of $h(X)$ and $A$. Denote $D^* = \underset{D}{\arg}\max~R_F(h;D)$. Then, for all $ s\in{\cal S}$ and $a \in {\cal A}$, 
\begin{align*}
    \dfrac{D^*(s,a)}{1-D^*(s,a)}  = \dfrac{p_{h(X),A}(s,a)}{p_{h(X)}(s)p_{A}(a)}=\dfrac{p_{h(X)|A}(s|a)}{p_{h(X)}(s)}.
\end{align*}
\end{proposition}
The proof is shown in Supplementary~\ref{pf:prop_sp}. This proposition provides a theoretical justification for the use of $R_F(h;D)$ as a GSP controller. Following the argument in Theorem~1 of \cite{good:etal:14}, we observe that $R_F(h)$ can be interpreted by the Jensen-Shannon divergence $J(\cdot,\cdot)$, i.e., $R_F(h;D^*)~=~2J(P(h(X),A),P(h(X))P(A)) - 2\log 2$, 
and it implies $p_{h(X),A}(s,a)=p_{h(X)}(s)p_{A}(a)$ for all $s$ and $a$ if $J=0$, which underpins that $h$ accomplishes GSP at the minimum of $R_F(h)$. Thus, we can formulate a fairness-aware optimization problem for $h$ by placing the extra penalty $R_F(h)$ to the (discriminatory) risk-minimization problem, i.e.,  
\begin{align}
\label{eqn:opt_gen}
    \underset{h}{\min}~ R(h) + \lambda R_{F}(h), 
\end{align}
where $\lambda$ trades off between goodness-of-fit (utility) and the degree of GSP; as $\lambda$ becomes larger, the solution model $h$ becomes fairer (in terms of GSP) where $D^*$ gets closer to 0.5 but meanwhile undergoes the loss of utility. 

Finally, we employ an empirical min-max optimization structure since \eqref{eqn:opt_gen} is practically intractable. The population joint density of $Y$, $X$, and $A$ is not available, and the optimal $D^*$ is unknown in general so it requires maximization over $R_F(h;D)$. Let's denote by $\{(X_i,A_i,Y_i)\}_{i=1}^n$ the observed data set with a sample size $n$. We collect $\{A_i'\}_{i=1}^n$ by simply resampling $\{A_i\}_{i=1}^n$. This is an approximated but computationally efficient implementation for sampling $A'$ from $P(A)$. Let's further denote by $\hat{R}(h)=\frac{1}{n}\sum_{i=1}^n L(Y_i,h(X_i))$ and $\hat{R}_F(h;D)=\frac{1}{n}\sum_{i=1}^n\left( \log D(h(X_i),A_i)  + \log (1-D(h(X_i),A'_i)) \right)$ the unbiased estimators of $R$ and $R_F$ respectively. Then, the empirical version of \eqref{eqn:opt_gen} is 
\begin{align}
\label{eqn:emp_opt}
    \min_h \max_D ~ \hat{R}(h)+\lambda \hat{R}_F(h;D),
\end{align}
where we use neural networks to model both $h$ and $D$. The optimization problem \eqref{eqn:emp_opt} can be solved via an alternative min-max strategy (Supplementary~\ref{appen:alg_gsp}). For every iteration, $D$ is first trained to capture the degree of discrimination against GSP. The model $h$ is then trained to minimize the risk plus the fairness penalty evaluated by $D$, which in turn hinders $D$ from identifying the presence of discrimination in the next iteration. By repeating this adversarial game between the two networks $D$ and $h$, the model $h$ settles down to an equilibrium, determined by $\lambda$, between utility and fairness. We name $\hat{R}_F(h;D)$ as \emph{simple random sampling (SRS)-based penalty (SBP)}.  

Although this work focuses on the Jensen-Shannon divergence, our main idea does readily apply to other metrics. For instance, a SBP penalty utilizing the Kantorovich-Rubinstein duality of the 1-Wasserstein distance can be devised as 
\begin{align*}
    R_F(h;C) &= \expect_{X,A} [C(h(X),A)] - \expect_{X,A'} [C(h(X),A')], 
\end{align*}
where $C$ belongs to the class of 1-Lipschitz functions \citep{arjo:etal:17}. In the same way, it is straightforward to see that other popular $f$-divergence metrics also have the specific formulation of $R_F$ \citep{nowo:etal:16}.

This penalty is not only adequately flexible to capture nonlinear discriminatory dependency between $h(X)$ and $A$ but also differentiable while most fairness criteria are usually non-differentiable in general \citep{zafa:etal:19,cott:etal:19}. Noteworthily, \emph{this penalty requires neither partitioning data sets into sub-groups nor predicting sensitive attributes} in contrast to a lot of previous works \citep{edwa:stor:15,xie:etal:17,beut:etal:17,zhan:etal:18,adel:etal:19,zhao:etal:19,li:etal:21,cho:etal:20,du:etal:21}. These properties greatly expand the applicability of the proposed method to numerous real-world problems having a mix of continuous and discrete sensitive attributes. We also remark that the previous sampler-based work of \cite{roma:etal:20} does not apply to GSP.

\subsection{Extension for Separation}
\label{sec:sep}
We also devise a penalty for separation that inherits all the advantages discussed in Section~\ref{sec:SBP}. To begin with, we define the generalized equalized odds (GEO) which is an intuitive generalization of equalized odds \citep{hard:etal:16} beyond binary outcomes. 
\begin{definition}[Generalized Equalized Odds (GEO)]
    The scoring model $h$ is called to satisfy the generalized equalized odds if $P(h(X)|A=a,Y=y)=P(h(X)|Y=y)$ for all $a\in {\cal A}$ and $y \in {\cal Y}$. 
\end{definition}
To control GEO, we suggest specifying $R_F(h;D)=$ 
\begin{align*}
 &\expect_{X,A,Y} [\log D(h(X),A,Y)] \\ 
 &+ \expect_{A'} \expect_{X,Y} [\beta(A',Y)\log (1-D(h(X),A',Y))], 
\end{align*}
with some function $\beta: \mathcal A\times \mathcal Y\rightarrow\mathbb R^{+}$ and $D:{\cal S}\times {\cal A} \times {\cal Y}\rightarrow (0,1)$, where $A'$ is statistically independent of $(X,A,Y)$ but $A$ and $A'$ have the same distribution. The proposed penalty accompanies the next proposition. 
\begin{proposition}
\label{prop:eo}
Let $p_{h(X)|Y}$ be the conditional density function of $h(X)$ given $Y$ and $p_{h(X)|A,Y}$ be of given $Y$ and $A$. For $R_F(h;D)$, if $D^* = \underset{D}{\arg}\max~R_F(h;D)$, then $D^*(s,a,y;\beta)=$  
\begin{align*}
    \dfrac{p_{h(X)|A,Y}(s|a,y)}{p_{h(X)|A,Y}(s|a,y)+ \beta(a,y) p_{h(X)|Y}(s|y)\frac{p_{A',Y}(a,y)}{p_{A,Y}(a,y)}},
\end{align*}
for all $s \in {\cal S}$, $a \in {\cal A}$, and $y \in {\cal Y}$, where $p_{A',Y}$ and $p_{A,Y}$ be the joint density functions of $A'$ and $Y$ and of $A$ and $Y$ respectively.
\end{proposition}

The proof appears in Supplementary~\ref{pf:prop_eo}. If $\beta$ cancels out the density ratio between $p_{A',Y}$ and $p_{A,Y}$ (i.e., $\beta(a,y)p_{A',Y}(a,y)/p_{A,Y}(a,y)=1$), then by the same argument in Proposition \ref{prop:sp}, $R_F(h;D^*)$ is minimized when $h$ satisfies $P(h(X)|A,Y)=P(h(X)|Y)$, i.e., the penalty $R_F(h;D^*)$ promotes GEO.
Thus, we fit a density-ratio estimator $\hat{\beta}$ by maximizing $R_{\beta}(D_{\beta})=$
\begin{align*}
    \expect_{A,Y} [\log D_{\beta}(A,Y)]+\expect_{A',Y}[\log (1-D_{\beta}(A',Y))],
\end{align*}
where $D_{\beta}:{\cal A}\times {\cal Y}\rightarrow (0,1)$ is modeled by a neural network.
Then $D_{\beta}^*=\arg_{D_{\beta}}\max~R_{\beta}(D_{\beta})$ satisfies
$D_{\beta}^*(a,y)/(1-D_{\beta}^*(a,y))=p_{A,Y}(a,y)/p_{A',Y}(a,y)$. Given an empirical minimizer $\hat D_{\beta}$ of $R_{\beta}(D_{\beta})$, let $\hat{\beta}(a,y)=\hat D_{\beta}(a,y)/(1-\hat D_{\beta}(a,y))$, and then by Proposition \ref{prop:eo}, $D^*(s,a,y;\hat{\beta})$ properly regularizes GEO. The numerical algorithm to expedite GEO appears in Supplementary~\ref{appen:alg} on the basis of the expression \eqref{eqn:emp_opt}. The algorithm needs a pre-training phase to estimate $\hat{\beta}$ that is leveraged as the adaptive weights for the evaluation of the penalty $R_F(h;D)$. We observe that $\hat{\beta}$ performs a powerful density-ratio estimation on toy examples in Supplementary~\ref{appen:geo_toy_exp}. It is worth mentioning that when $A$ and $Y$ are both discrete, $\hat{\beta}$ can be easily found without neural network training, e.g., using the empirical probability mass functions.

Proposition~\ref{prop:eo} clearly shows the fundamental differences of our simple random sampler approach compared to the conditional sampler approach \citep{roma:etal:20}, i.e., $A' \sim P(A|Y)$. In our case, we additionally employ $\beta$ to exactly match $P(h(X)|A,Y)=P(h(X)|Y)$ in the ideal separation. In contrast, the prior work does not need $\beta$ but has to obtain the conditional sampler $A'=G(\epsilon,Y)$ for some generative model $G$ and random noise $\epsilon$. There may be room for debate, but we believe that estimating $\beta$ is generally more straightforward than training $G$. The former is equivalent to a DNN binary classification problem, which in modern practice is hardly impacted by the dimension or characteristics of $A$ for tabular data sets. In contrast, the latter may necessitate the use of advanced generative models, such as generative adversarial networks (GAN) or diffusion models which are well-known for their notorious training difficulty \citep{xule:etal:19,kote:etal:23}, when dealing with tabular-type attributes of $A$. Moreover, even a successfully trained generative model may only learn the support rather than the shape of a distribution \citep{aror:etal:18}.

\begin{remark}
    Although our approach necessitates $\beta$ for the theoretical justification, we find that the performance of our simple random sampler-based method is fairly robust to a poor $\beta$ estimation. Refer to Section~\ref{sec:sim} and Supplementary~\ref{sec:robust_beta} for more details.
\end{remark}

Finally, it is worth mentioning that this penalty can also be used to obtain fair representation \citep{zhao:etal:19,du:etal:21}. Let's denote by $E:{\cal X} \rightarrow {\cal E}$ an encoder which precedes $h_E:{\cal E}\rightarrow {\cal S}$ with the risk $\hat{R}(h_E \circ E)$. Then solving $\min_{h_E,E} \max_D \hat{R}(h_E \circ E)+\lambda \hat{R}_F(E;D)$ brings a fair encoder $E$ for either independence or separation with $D:{\cal E}\times {\cal A}\rightarrow (0,1)$ or $D:{\cal E}\times {\cal A}\times {\cal Y}\rightarrow (0,1)$ respectively.

\section{THEORY}
\label{sec:theory}

In this section, we characterize the estimation error and the loss of utility of the proposed fairness-aware optimization scheme. For the simplicity of presentation, the analysis focuses on the solution of \eqref{eqn:emp_opt} with the GSP penalty. Our analysis borrows some proof techniques in the literature of generative adversarial modeling \citep{ji:etal:21}. For readers who are interested in the details of the proof, please refer to Supplementary~\ref{appen:theory}. 

Let's suppose ${\cal X}=\{x:||x||\leq B, \, x\in \mathbb{R}^p\}$, ${\cal Y}=[0,1]$, and ${\cal A}=[0,1]^l$ where $||\cdot||$ denotes the Euclidean norm. For mathematical convenience, we set $D(\cdot)=\sigma(f(\cdot))$ with $\sigma(x)=(1+\exp(-x))^{-1}$ and consider the fully-connected neural networks: $f(x,a)=f_{\bw}(x,a)=w^{\top}_d \kappa_{d-1}(W_{d-1}\kappa_{d-2}(\cdots W_1[x^{\top},a^{\top}]^{\top}))$ and $h(x)=h_{\bv}(x)=v^{\top}_g \psi_{g-1}(V_{g-1}\psi_{g-2}(\cdots V_1x))$ where $\bw=(W_1,\dots, W_{d-1}, w_d)\in \bW$ and $\bv=(V_1,\dots, V_{g-1}, v_g)\in \bV$. We denote by ${\cal F}$ and ${\cal H}$ the function classes of $f$ and $h$. Also, it is assumed that 
$\bW = \bigotimes_{i=1}^{d-1}\{W_i \in \mathbb{R}^{p_{i+1} \times p_{i}} : ||W_i||_F \leq M_w(i) \} \bigotimes \{w_d \in \mathbb{R}^{p_d \times 1} : ||w_d|| \leq M_w(d)\}$
and $\bV = \bigotimes_{i=1}^{g-1}\{V_i \in \mathbb{R}^{q_{i+1} \times q_{i}}: ||V_i||_F \leq M_v(i) \} \bigotimes \{v_g \in \mathbb{R}^{q_g\times 1}: ||v_g|| \leq M_v(g)\}$ with constants $M_w(\cdot)$ and $M_v(\cdot)$, $p_1=p+l$, $q_1=p$, and the Frobenius norm $||\cdot||_F$. This also induces the class of $D$ by functional composition, defined as ${\cal D}=\{f_{\bw}(h_{\bv}(x),a): \bw \in \bW,  \bv \in \bV\}$.

We further assume that the activation functions $\psi_u$ and $\kappa_t$ are $K_{\psi}(u)$ and $K_{\kappa}(t)$-Lipschitz for all $t=1,\dots,d-1$ and $u=1,\dots,g-1$. ReLU and Sigmoid are examples of the 1-Lipschitz functions. These assumptions hint $1 > \gamma_1 \geq \sigma(h_{\bv}(x))\geq \gamma_0 > 0$ and $1 > \nu_1 \geq \sigma(f_{\bw}(x,a)) \geq \nu_0 > 0$ for all $x$ and $a$ where the upper and the lower limits $(\gamma_1,\gamma_0,\nu_1,\nu_0)$ depend on the assumed bounds from the parameter spaces (i.e., $M_w(\cdot)$ and $M_v(\cdot)$) and the Lipschitz constants of the activation functions.

Now, let's denote $d(h;\lambda)=R(h) + \lambda R_F(h)$ and $\hat{d}(h;\lambda)=\hat{R}(h) + \lambda \hat{R}_F(h)$ as 
the population-level target function and its empirical version respectively. The estimation error then can be characterized by evaluating the empirical solution on the population objective based on the Rademacher complexity. 
\begin{definition}[Rademacher Complexity]
    Let ${\cal H}$ be the function class of $h$. Denote by $X_1,\dots,X_n$ random samples that are independent and identically distributed (i.i.d.) to $P_X$. Then the Rademacher complexity ${\cal R}({\cal H})$ is defined as 
    \begin{align*}
        {\cal R}({\cal H})=\expect_{X, \epsilon} \left[\sup_{h \in {\cal H}} \left|\dfrac{1}{n}\sum_{i=1}^n \epsilon_i h(X_i) \right| \right],
    \end{align*}
    where $\epsilon_1,\dots,\epsilon_n \sim {\rm Unif}\{-1,1\}$ i.i.d.
\end{definition}
The Rademacher complexity for other function classes is defined in the same fashion. 
Theorem~\ref{thm:ee} holds for either the cross-entropy loss with $Y\in\{0,1\}$ or the mean absolute error for $Y\in [0,1]$. 
\begin{theorem}[Estimation Error]
\label{thm:ee}
    Let $\hat{h}^* = \arg_{h\in {\cal H}}\min \hat{d}(h;\lambda)$ and define ${\cal L}=\{L(y,h_{\bv}(x)):\bv \in V\}$ for $x\in {\cal X}$ and $y\in {\cal Y}$. Then, for any given $\lambda \geq 0$, an upper bound of the estimation error is $|d(\hat{h}^*;\lambda)-\inf_{h \in {\cal H}} d(h;\lambda)| \leq$
    \begin{align*}
    & \underbrace{4{\cal R}({\cal L}) + 2F_{\bV,\psi,B,\gamma_0,\gamma_1}\sqrt{\dfrac{\log(1/\delta)}{2n}}}_{\text{from }\hat{R}(h)} \\
    &+ \underbrace{2\lambda \left(F_{\nu_0,\nu_1}{\cal R}({\cal D}) + F_{\bW,\bV,B,\kappa,\psi,l,\nu_0,\nu_1}\sqrt{\dfrac{\log(1/\delta)}{2n}}\right)}_{\text{from } \hat{R}_F(h)},
    \end{align*}
    with the probability $1-3\delta$ where $F_{\nu_0,\nu_1}$, $F_{\bV,\psi,B,\gamma_0,\gamma_1}$, and $F_{\bW,\bV,B,\kappa,\psi,l,\nu_0,\nu_1}$ are constants depending on the architectures of neural networks $h\in{\cal H}$ and $f\in{\cal F}$ whose exact values can be found in Supplementary~\ref{appen:theory}.  
\end{theorem}
Theorem \ref{thm:ee} shows how the complexity of the neural network function classes affects the estimation accuracy through the Rademacher complexity and $F_{\cdot}$ constants. As sample size $n$ increases (while other settings are fixed), the Rademacher complexity generally decreases to zero, leading to consistency according to Theorem \ref{thm:ee}.
The hyperparameter $\lambda$ controls the balance between utility and fairness. Larger $\lambda$ enforces better fairness at the expense of possible utility loss; the next corollary studies this utility loss.




\begin{cor} [Loss of Utility]
\label{thm:cor}
 Define $\hat{h}^* = \arg_{h\in{\cal H}}\min \hat{d}(h;\lambda)$,  
 ${h}^*_0 = \arg_{h\in{\cal H}}\min  {d}(h;\lambda=0)$, and ${h}^* = \arg_{h\in{\cal H}}\min  {d}(h;\lambda)$. Let's denote $\Delta(h_0^*, h^*)=R_F(h_0^*)-R_F(h^*)$. Then the loss of utility has an upper bound $|d(\hat{h}^*;\lambda=0)-d(h^*_0;\lambda=0)|\leq $ 
\begin{align*}
 4{\cal R}({\cal L}) + 2F_{\bV,\psi,B,\gamma_0,\gamma_1}\sqrt{\dfrac{\log(1/\delta)}{2n}}+\lambda \Delta(h_0^*, h^*),
\end{align*}
with 1-$\delta$ probability for any $\lambda \geq 0$.
\end{cor}
The loss of utility can take place from two main sources. Basically, the loss of utility could stem from sampling errors, which unfolds the first two terms in the upper bound of Corollary~\ref{thm:cor}. Secondly, if the true $h_0^*$, which is of the best utility, is exposed to huge discrimination against $A$ in the population, i.e., $h$ statistically strongly depends on $A$, then the penalty could lead to tremendous loss of utility in an effort to remove the dependency w.r.t. $A$ (i.e., the term $\lambda \Delta(h_0^*, h^*)$). 

\section{SIMULATION}
\label{sec:sim}

The performance of supervised learning with the proposed penalties is verified in the following three scenarios: (I) discrete outcome and sensitive attribute, (II) discrete outcome and mixed sensitive attributes, and (III) continuous outcome and sensitive attribute. The main text only delivers results about GSP in Scenario I and GEO in Scenario II due to the page limit. Results about GEO in Scenario I,  GSP in Scenario II, Scenario III, and more additional studies (e.g., fair representation) are presented in Supplementary Section~\ref{appen:sim}.

We compare the performance of our model (SRS-based penalty, SBP) against several newly proposed competing methods: HGR \citep{lee:etal:22} that uses the neural-net approximation to calculate the soft HGR, the feature neutralization (\cite{du:etal:21}, NEU) that interpolates feature points in a hidden layer such that the mapped space is independent of sensitive attributes, the kernel density estimation (\cite{cho:etal:20}, KDE) that employs $Q$-function to approximate a distribution function, and CON \citep{roma:etal:20} that uses the learned conditional sampler of $A$ given $Y$. All the above in-processing models have the trade-off parameter $\lambda$ to the fairness penalty $L_F$ for fairness on top of the main loss term $L_M$. 
As a means of matching the scale of $\lambda$ for the different models, all of them are tested under the formula $(1-\lambda)L_M+\lambda L_F$ with 5 replicated experiments for various $0<\lambda<1$. All methods learn a neural network $h$ with 3 hidden layers with 64 nodes. For more details about simulation setups, please refer to Supplementary Section~\ref{appen:sim_setting}. 

Three benchmark data sets are considered for Scenario I: {\bf Adult Data}\footnote{\label{data:uci}https://archive.ics.uci.edu/ml/datasets/} where $Y$ is whether or not the annual income is greater than \$50K and $A$ is whether an individual is white or non-white; {\bf Law School Admission Data }\footnote{http://www.seaphe.org/databases.php} where $Y$ is whether or not an applicant receives admission and $A$ whether an individual is white or non-white; and {\bf Credit Card Default Data}\footref{data:uci} where $Y$ is whether or not a customer declares default and $A$ is the gender. We refer to the work \citep{cho:etal:20} to specify the sensitive attributes for analysis. Each data set is split by 80\% and 20\% for training and validation during the training course. 

For the evaluation of fairness in Scenario I, statistical parity (SP) is assessed by calculating 
\begin{align*}
    \text{SP} &=\left| \expect (\hat{Y}=1|A=1) / \expect (\hat{Y}=1|A=0) -1 \right|, 
\end{align*}
where ${\cal Y}=\{0,1\}$, ${\cal A}=\{0,1\}$, and  $\hat{Y}=I(\hat{h}^*(X)>\tau)$ with the threshold $\tau$ maximizing the area under ROC curve (AUC) on the validation set. The AUC value is also used as a measure of utility. We use the Kolmogorov-Smirnov statistics (KS) to assess GSP, which is calculated as 
\begin{align*}
    \text{KS-GSP} &= \sum_{a\in {\cal A} }\max_{h_x} |\hat{P}(\hat{h}^*(X)\leq h_x | A=a) \\
     &~~~~~~~~~~~~~~~~~~~~~~~~~~~~~~-\hat{P}(\hat{h}^*(X)\leq  h_x)|,
\end{align*}
where $\hat{P}$ denotes the empirical distribution.

\begin{figure*}[hbt!]
    \centering
    \includegraphics[trim={2.5cm 0 0 0},clip,scale=0.40]{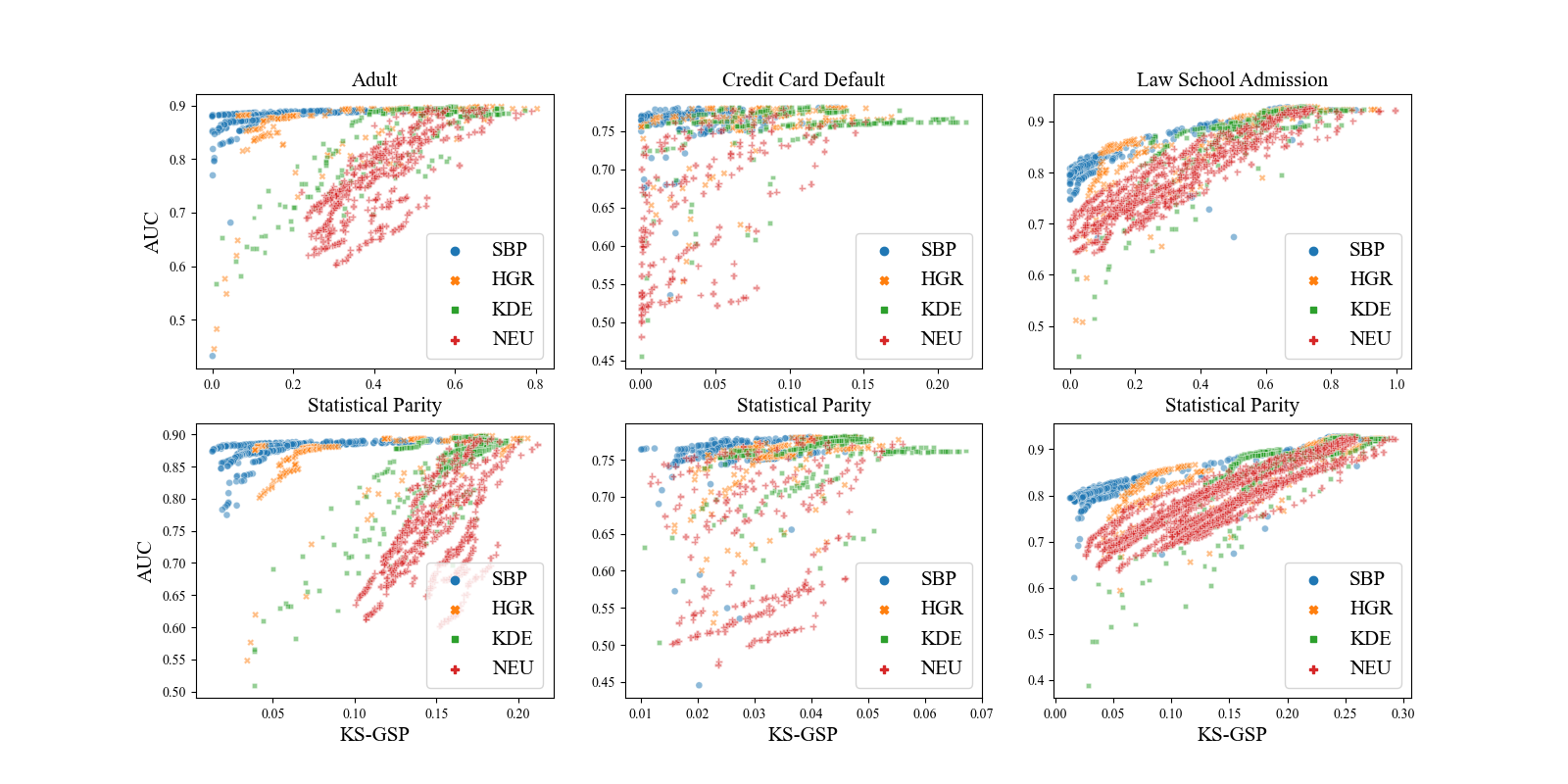}
    \caption{(Scenario I) Pareto frontiers: the first row includes pairs of SP and AUC, and the second row shows pairs of KS-GSP and AUC from 5 experiments for each $\lambda$. SBP (ours) tends to be more tightly in the upper-left corner than the competitors.}
    \label{sim:s1sp}
\end{figure*}

Since the model intrinsically has multi-objectives (utility and fairness) in nature, the trained models at all iterations are evaluated over the validation data. Then, we plot the Pareto frontier curve of (AUC versus SP/KS) for each independent run. The Pareto frontier \citep{emme:etal:18} is a set of solutions that are not dominated by other pairs. For instance, a pair of AUC and SP, e.g., $(0.5,0.5)$, is not dominated by $(0.4,0.4)$ but by $(0.6,0.4)$. This analysis helps design fair comparison studies without designing different or non-comparable early stopping strategies for the competitors (for example, NEU tends to degenerate to the trivial solution, which is perfectly fair but has no utility, for long-run iterations, so comparing the last iterates among the competing methods is not fair).

\begin{table*}[hbt!]
\caption{Averages of the 5 smallest SP/KS-GSPs whose AUCs are greater than the thresholds. Those scores are selected in the Pareto solutions appearing in Figure~\ref{sim:s1sp}. Standard deviations are in the parentheses next to the averages. All values are rounded to the third decimal place.}
\label{tab:sce1_sp1}
\centering
\begin{tabular}{c|cc|cc|cc}
\toprule
& \multicolumn{2}{c|}{\textbf{Adult} (AUC $\geq$ 0.85)} & \multicolumn{2}{|c|}{\textbf{Cred. Card.} (AUC $\geq$ 0.75)} & \multicolumn{2}{|c}{\textbf{Law School.} (AUC $\geq$ 0.80)} \\
\hline
Model & SP   ($\downarrow$)    & KS-GSP  ($\downarrow$)   & SP   ($\downarrow$)        & KS-GSP     ($\downarrow$)     & SP    ($\downarrow$)         & KS-GSP  ($\downarrow$)      \\
\hline
\hline
SBP  & 0.001 ($\approx 0$) & 0.014 ($\approx 0$) &  $\approx 0$ ($\approx 0)$   & 0.011 (0.001)       & 0.004 (0.003)       & 0.018 (0.001)       \\
HGR  & 0.070 (0.005)       & 0.040 ($\approx 0$) &  0.001 (0.001)               & 0.023 (0.001)       & 0.096 (0.003)       & 0.064 (0.002)       \\
KDE  & 0.355 (0.010)       & 0.124 (0.002)       &  0.013 (0.009)               & 0.024 ($\approx 0)$ & 0.239 (0.013)       & 0.132 (0.003)       \\
NEU  & 0.446 (0.029)       & 0.154 (0.003)       &  0.040 (0.010)               & 0.020 (0.003)       & 0.214 (0.017)       & 0.098 (0.005)       \\ 
\bottomrule   
\end{tabular}
\end{table*}


Figure~\ref{sim:s1sp} clearly illustrates the remarkable advantages of using SBP over the competitors. The closer the set of points is to the upper left corner of the figure, the higher performance is verified.  We observe that SBP provides more consistent and better solutions in the sense that the Pareto frontiers are much more tightly gathered along the trade-off path than other competing methods. Note CON is not available for SP. Table~\ref{tab:sce1_sp1} reports the fairness scores (SP/KS) of the Pareto solutions whose AUC is above the thresholds. The table succinctly demonstrates our superiority which is consistent with Figure~\ref{sim:s1sp}. Those thresholds are chosen such that all methods yield a sufficient number of candidate SP/KS scores by referring to Figure~\ref{sim:s1sp}. More tables with different thresholds appear in Supplementary~\ref{appen:tabs}.

\begin{table*}[hbt!]
\centering
\caption{Averages of the 5 smallest EO/KS-GEOs whose AUCs are greater than the thresholds. Those scores are selected by referring to Figure~\ref{sim:s2eo} in Supplementary~\ref{appen:simul_sp}. Standard deviations are in the parentheses.}
\label{tab:s2eo}
\begin{tabular}{c|c|cc|cc|cc}
\toprule
 &  & \multicolumn{2}{c|}{\textbf{Adult} (AUC $\geq$ 0.80)} & \multicolumn{2}{|c|}{\textbf{Cred. Card.} (AUC $\geq$ 0.75)} & \multicolumn{2}{|c}{\textbf{ACSEmpl.} (AUC $\geq$ 0.75)} \\
\hline
 Metric & Model & Race  ($\downarrow$)    &     Age   ($\downarrow$)   &    Gender   ($\downarrow$)    &   Age  ($\downarrow$)     &    Gender   ($\downarrow$)      &  Age     ($\downarrow$)     \\
\hline
\hline
\multirow{4}{*}{EO}& SBP  &  0.047 (0.004)   &  0.078 (0.002)   &  0.001 ($\approx$ 0)   &  0.016 (0.001)    &   0.066 (0.004)      &   0.209 (0.002)      \\
                 & CON &  0.358 (0.029) &  0.162 (0.027)      &  0.006 (0.003)         &  0.016 (0.004)    &  0.155 (0.006)      &  0.316 (0.011)   \\
                 & HGR  &  0.273 (0.073)   &  0.307 (0.002)   &  0.002 (0.001)      &   0.033 (0.001)     &   0.193 (0.005)       &  0.447 (0.002)         \\
                 & NEU  &  0.292 (0.019)   &  0.201 (0.007)   &  0.027 (0.007)      &   0.055 (0.005)     &   0.099 (0.003)       &  0.276 (0.006)      \\ 
\hline
\hline
\multirow{4}{*}{\begin{tabular}[c]{@{}l@{}}KS-\\GEO\end{tabular}}& SBP  & 0.098 (0.003)  &  0.120 (0.001)  &  0.040 (0.001)    &  0.036 (0.001)     &  0.071 (0.001)       &  0.167 (0.002)       \\
                    & CON   & 0.177 (0.001) &  0.094 (0.007)  &  0.049 (0.001)   &  0.038 ($\approx 0$) & 0.100 (0.002)       &  0.207 (0.003)     \\
                    & HGR  & 0.160 (0.003)  &  0.208 (0.008)  &  0.041 (0.003)    &  0.047 (0.001)     &  0.122 (0.001)       &  0.320 (0.004)       \\
                    & NEU  & 0.156 (0.005)  &  0.130 (0.001)  &  0.063 (0.004)    &  0.054 (0.001)     &  0.074 (0.001)       &  0.205 (0.002)       \\ 
\bottomrule
\end{tabular}
\end{table*}

For Scenario II, we use an additional data set {\bf ACSEmployment}\footnote{https://github.com/socialfoundations/folktables} from California in 2018 where $Y$ is whether or not an individual is employed and $A$ is a vector of age (continuous) and gender (discrete). We choose age (continuous) and race (discrete) for Adult and also age (continuous) and gender (discrete) for Credit Card Default as sensitive attributes respectively while the output variables are the same as in Scenario I. For CON, we devise a conditional GAN model \citep{mirz:osin:14} because the trivial estimation method for $P(A|Y)$ in the original work \citep{roma:etal:20} is not directly applicable when $A$ is a mix of discrete and categorical variable. For a fair comparison, we use the same network structure for $\beta$ in SBP and the discriminator for CON. To measure fairness for discrete attributes, we define 
\begin{align*}
\text{EO}=\sum_{y\in {\cal Y}}\left|\frac{\expect (\hat{Y}=1|A=1,Y=y)}{\expect(\hat{Y}=1|A=0,Y=y)}-1\right|,
\end{align*}
and $\text{KS-GEO}=$
\begin{align*}
\sum_{y\in{\cal Y},a\in{\cal A}} \max_{h_x} |&\hat{P}(\hat{h}^*(X)\leq h_x | A=a, Y=y)\\
&-\hat{P}(\hat{h}^*(X)\leq h_x |Y=y)|.    
\end{align*}
We similarly define the fairness measures for continuous attributes; please refer to  Supplementary~\ref{appen:simul_sp} to see the formal definition.

\begin{table}[hbt!]
\centering
\caption{Training times (mins) for the first 1000 iterations on A30 GPU for separation in \textbf{Adult}, including/excluding the pre-training time. Note that for Scenario II, CON, NEU, and SBP need a pre-training (additional NN models) course but KDE and HGR do not. Refer to Supplementary~\ref{appen:sim_setting} for more details.}
\begin{tabular}{c|cc|c}
\toprule
 & \multicolumn{2}{c|}{Without}& \multicolumn{1}{c}{With } \\
  & \multicolumn{2}{c|}{pre-training}& \multicolumn{1}{c}{pre-training} \\ 
 \hline
\multicolumn{1}{c|}{Model} &  Scenario I  &  Scenario II    &     Scenario II \\ 
 \hline
 \hline
\multicolumn{1}{c|}{SBP} &  0.22 (0.02)   & 0.59 (0.05)   &  0.80 (0.07)     \\
\multicolumn{1}{c|}{CON} &  0.21 (0.03)   & 0.40 (0.03)   &  1.45 (0.09)          \\
\multicolumn{1}{c|}{HGR} &  0.47 (0.08)   & 0.77 (0.10)   & -  \\
\multicolumn{1}{c|}{KDE} &  1.31 (0.09)   & -   &           -  \\
\multicolumn{1}{c|}{NEU} &  1.01 (0.09)   & 1.46 (0.03)   &  3.61 (0.09)  \\
\bottomrule
\end{tabular}
\label{tab:tr_time}
\end{table}


Table~\ref{tab:s2eo} contrasts all models except KDE on the three data sets. Note KDE can not be implemented for continuous sensitive attributes. The thresholds are chosen based on Figure~\ref{sim:s2eo} in Supplementary~\ref{appen:simul_sp}. As is consistent with the table, the figure illustrates that SBP tends to outperform the competitors for both sensitive variables in general. For the computing time, ours is generally more efficient as shown in Table~\ref{tab:tr_time}. 

To intuitively explain that SBP outperforms others, we comment that HGR involves more burdensome approximation because it has to estimate four auxiliary neural networks in the adversarial training procedure, leading to larger estimation errors and computing time; NEU, on the other hand, does not explicitly control fairness metrics in its training process, leading to uneven performance. CON is significantly defeated by SBP mainly because estimating a good conditional generator $A|Y$ via the GAN approach is more difficult than estimating a good $\hat\beta$. On the other hand, we observe that SBP for separation is robust against a poor estimation of $\hat{\beta}$ (details in Supplementary~\ref{sec:robust_beta}).

In spite of such remarkable performance, an unavoidable issue of the proposed penalty remains regarding the adversarial game structure between $h$ and $D$ which is known to be unstable in general. As remedies to this obstacle, users can harness popular stabilization techniques used in the GAN literature such as penalizing the gradient of $D$ \citep{zhou:etal:19} or adding normalization layers \citep{miya:etal:18}. It is also practical for the users to save the snapshots of $h$ being learned over the Pareto frontier and deploy the final model having the desired Pareto solution for the purpose.

\section{CONCLUSION}
The proposed fair-ML framework that employs the simple random sampler is universally applicable to classification/regression problems across various fairness criteria, without worrying whether sensitive attributes are continuous or discrete, or even a mix of them. This methodological versatility is of great importance as multifarious communities in society pay more attention to the unbiased data-driven decision process while having more and more diverse variables with distinctive characteristics to be protected together \citep{louk:etal:19,lee:flor:21,kozo:etal:22,dear:etal:22}. 
In future research, the fundamental concept of unconditionally sampling sensitive attributes will be a promising tool for promoting fairness for other related works that heavily depend on the discreteness, mixed or even high-dimensional sensitive variables, such as synthesizing a fair data set \citep{xu:etal:19} or fair uncertainty quantification \citep{lu:etal:22}.

\section*{Acknowledgements}
G. Lin acknowledges the support of the National Science Foundation (DMS-2053746, DMS-2134209, ECCS-2328241, and OAC-2311848), and U.S. Department of Energy (DOE) Office of Science Advanced Scientific Computing Research program DE-SC0023161, and DOE–Fusion Energy Science, under grant number: DE-SC0024583.


\bibliography{sample}




\section*{Checklist}


 \begin{enumerate}
 
 \item For all models and algorithms presented, check if you include:
 \begin{enumerate}
   \item A clear description of the mathematical setting, assumptions, algorithm, and/or model. [\textbf{Yes}/No/Not Applicable] $\rightarrow$ Refer to  Sections~\ref{sec:SBP} and \ref{sec:sep}.
   \item An analysis of the properties and complexity (time, space, sample size) of any algorithm. [\textbf{Yes}/No/Not Applicable]$\rightarrow$ See Table~\ref{tab:tr_time}.
   \item (Optional) Anonymized source code, with specification of all dependencies, including external libraries. [\textbf{Yes}/No/Not Applicable] $\rightarrow$ Refer to \href{https://github.com/jwsohn612/fairSBP}{the author's Github}. The code source was anonymized during the review process.
 \end{enumerate}

 \item For any theoretical claim, check if you include:
 \begin{enumerate}
   \item Statements of the full set of assumptions of all theoretical results. [\textbf{Yes}/No/Not Applicable] $\rightarrow$ Refer to Section~\ref{sec:theory}.
   \item Complete proofs of all theoretical results. [\textbf{Yes}/No/Not Applicable] $\rightarrow$ Refer to Supplementary~\ref{appen:theory}.
   \item Clear explanations of any assumptions. [\textbf{Yes}/No/Not Applicable] $\rightarrow$ Refer to Section~\ref{sec:theory}.
 \end{enumerate}

 \item For all figures and tables that present empirical results, check if you include:
 \begin{enumerate}
   \item The code, data, and instructions needed to reproduce the main experimental results (either in the supplemental material or as a URL). [\textbf{Yes}/No/Not Applicable] $\rightarrow$ Refer to \href{https://github.com/jwsohn612/fairSBP}{the author's Github}.
   \item All the training details (e.g., data splits, hyperparameters, how they were chosen). [\textbf{Yes}/No/Not Applicable] $\rightarrow$ Refer to Supplementary~\ref{appen:sim_setting} and Section~\ref{sec:sim}. 
   
 \item A clear definition of the specific measure or statistics and error bars (e.g., with respect to the random seed after running experiments multiple times). [\textbf{Yes}/No/Not Applicable] $\rightarrow$ Refer to Section~\ref{sec:sim} and Supplementary~\ref{appen:sim}.
 \item A description of the computing infrastructure used. (e.g., type of GPUs, internal cluster, or cloud provider). [\textbf{Yes}/No/Not Applicable] $\rightarrow$ Refer to Table~\ref{tab:tr_time}.
 \end{enumerate}

 \item If you are using existing assets (e.g., code, data, models) or curating/releasing new assets, check if you include:
 \begin{enumerate}
   \item Citations of the creator If your work uses existing assets. [\textbf{Yes}/No/Not Applicable] $\rightarrow$ Refer to Section~\ref{sec:sim}. Competing methods are cited.
   \item The license information of the assets, if applicable. [Yes/No/\textbf{Not Applicable}]
   \item New assets either in the supplemental material or as a URL, if applicable. [\textbf{Yes}/No/Not Applicable] $\rightarrow$ Refer to \href{https://github.com/jwsohn612/fairSBP}{the author's Github}.
   \item Information about consent from data providers/curators. [Yes/No/\textbf{Not Applicable}]
   \item Discussion of sensible content if applicable, e.g., personally identifiable information or offensive content. [Yes/No/\textbf{Not Applicable}]
 \end{enumerate}

 \item If you used crowdsourcing or conducted research with human subjects, check if you include:
 \begin{enumerate}
   \item The full text of instructions given to participants and screenshots. [Yes/No/\textbf{Not Applicable}]
   \item Descriptions of potential participant risks, with links to Institutional Review Board (IRB) approvals if applicable. [Yes/No/\textbf{Not Applicable}]
   \item The estimated hourly wage paid to participants and the total amount spent on participant compensation. [Yes/No/\textbf{Not Applicable}]
 \end{enumerate}

 \end{enumerate}

\onecolumn
\aistatstitle{Supplementary Materials: Fair Supervised Learning with A Simple Random Sampler of Sensitive Attributes}

\section{DETAILS OF SECTION 3.1}
\subsection{Proof of Proposition~\ref{prop:sp}} 
\label{pf:prop_sp}
Let's recall 
\begin{align*}
    R_F(h;D)=\expect_{X,A} [\log D(h(X),A)] + \expect_{X,A'} [\log (1-D(h(X),A'))],
\end{align*}
where $A$ and $A'$ are independent but identically distributed.
Let $s=h(x)$ where $x$ is a realization of $X$ and also $a$ is of $A$.  The loss function can be written 
\begin{align*}
    R_F(h;D) &= \int \log D(s,a)p(s,a)dsda + \int \log (1-D(s,a'))p(s,a')dsda', \\
    &= \int \log D(s,a)p(s|a)p(a) +  \log (1-D(s,a))p(s)p(a) dsda. 
\end{align*}
By the proof of Proposition 1 in \cite{good:etal:14},  $R_F(h;D)$ is maximized at \begin{align*}
    D^*(s,a)=\dfrac{p(s|a)p(a)}{p(s|a)p(a)+p(s)p(a)}=\dfrac{p(s|a)}{p(s|a)+p(s)},
\end{align*}
for any $s\in {\cal S}$ and $a\in{\cal A}$. 

\subsection{Algorithm of GSP penalty}
\label{appen:alg_gsp}

Let $\{(x_i,a_i,y_i)\}_{i=1}^n$ be a set of the realization of $\{X_i,A_i,Y_i\}_{i=1}^n$. The simple random sampler of $A$ can be easily obtained by the minibatch construction (Algorithm~\ref{alg:minibatch}). In practice, producing $a_{(i)}'$ directly from ${\bf D}$, which is an approximated implementation of Algorithm~\ref{alg:minibatch}, works well. In every iteration, $D$ serves as a fairness critic by quantifying the degree of discrimination against GSP. The model $h$ is then trained to minimize the risk but at the same time to be debiased such that $D$ would not capture the discrimination. Note $\bw$ and $\bv$ are parameters of $D$ and $h$. The algorithm to control GSP appears in Algorithm~\ref{alg:gsp}. It is allowed to update $\bw$ up to $T'\geq 1$ times for every single update of $\bv$ for better approximation for $\hat{R}_F(h)$, e.g., $T'=1,\dots,10$. 

\begin{algorithm}[th]
\caption{Minibatch Construction (MC) at the $t$th iteration}\label{alg:minibatch}
\KwData{Let ${\bf D}_n=\{(x_i,a_i,y_i)\}_{i=1}^n$ be the set of training data set. Set the minibatch size $n_b$. The subscript $(i)$ denotes the $i$th drawn sample.}
\KwResult{${\bf D}_{n_b}$}
    ${\bf D}=\{(x_{(i)},a_{(i)},y_{(i)})\}_{i=1}^{n_b}$ is randomly drawn from ${\bf D}_n$. \\ 
    ${\bf D}'=\{(x_{(i)}',a_{(i)}',y_{(i)}')\}_{i=1}^{n_b}$ is randomly drawn from ${\bf D}_n\setminus {\bf D}$. \\
    Construct ${\bf D}_{n_b}=\{(x_{(i)},a_{(i)},y_{(i)},a_{(i)}')\}_{i=1}^{n_b}$ by selecting $\{a_{(i)}'\}_{i=1}^{n_b}$ from ${\bf D}'$ and combining it into ${\bf D}$.
\end{algorithm}

\begin{algorithm}[th]
\caption{Generalized Statistical Parity}\label{alg:gsp}
\KwData{Let ${\bf D}_n=\{(x_i,a_i,y_i)\}_{i=1}^n$ be the set of training data set. Denote the $t$th iterate of model parameters by $\bw^{(t)}$  and $\bv^{(t)}$ and $\alpha$ is a learning rate. Fix the number of training iterations $T$ and $T'$; set $t=0$; and initialize $\bw^{(0)}$ and $\bv^{(0)}$.}
\KwResult{$h_{\bv^{(T)}}$}
\While{$t \leq T$}{
    $t=t+1$\\ 
    
    $\{(x_{(i)},a_{(i)},y_{(i)},a_{(i)}')\}_{i=1}^{n_b} = \text{MC}({\bf D}_n)$ \\
    
    $t' = 0$\\ 
    \While{$t' \leq T'$}{
    
    $t' = t' + 1$ \\ 

    $\hat{R}_F(h_{\bv};D_{\bw^{(t)}})=\frac{1}{n_b} \sum_{i=1}^{n_b} \left( \log D_{\bw^{(t)}}(h_{\bv^{(t)}}(x_{(i)}),a_{(i)})  + \log (1-D_{\bw^{(t)}}(h_{\bv^{(t)}}(x_{(i)}),a'_{(i)})) \right)$\\
    
    $\bw^{(t+1)} = \bw^{(t)} + \alpha \dfrac{\partial}{\partial \bw^{(t)}}\hat{R}_F(h_{\bv^{(t)}};D_{\bw^{(t)}})$\\ 
    }

    $\hat{R}(h_{\bv^{(t)}})=\frac{1}{n_b}\sum_{i=1}^{n_b} L(y_{(i)}, h_{\bv^{(t)}}(x_{(i)}))$ \\

    $\bv^{(t+1)} = \bv^{(t)} - \alpha \dfrac{\partial}{\partial \bv^{(t)}} (\hat{R}(h_{\bv^{(t)}}) +  \lambda \hat{R}_F(h_{\bv^{(t)}};D_{\bw^{(t)}}))$ \\ 
}
\end{algorithm}


    

    



\newpage

\section{DETAILS OF SECTION~\ref{sec:sep}}

\subsection{Proof of Proposition~\ref{prop:eo}}
\label{pf:prop_eo}
The penalty for separation is 
\begin{align*}
    R_F(h;D)&=\expect_{X,A,Y} [\log D(h(X),A,Y)] +  \expect_{A'} \expect_{X,Y} [\beta(A',Y) \log (1-D(h(X),A',Y))], \\
    &=\int \log D(s,a,y)p(s|a,y)p(a,y)dadyds + \beta(a',y)\log (1-D(s,a',y))p(s|y)p(a')p(y)da'dyds,\\ 
    &=\int \log D(s,a,y)p(s|a,y)p(a,y) + \beta(a,y)\log (1-D(s,a,y))p(s|y)p(a)p(y)dadyds.
\end{align*}
By the same argument in Supplementary~\ref{pf:prop_sp}, 
$D^*=\arg_D \max R_F(h:D)$ has the form of 
\begin{align*}
    D^*(s,a,y;\beta) = \dfrac{p(s|a,y)p(a,y)}{p(s|a,y)p(a,y)+\beta(a,y) p(s|y)p(a)p(y)} =\dfrac{p(s|a,y)}{p(s|a,y)+\beta(a,y) p(s|y) \dfrac{p(a)p(y)}{p(a,y)}}.
\end{align*}

\subsection{Algorithm of GEO penalty}
\label{appen:alg}

GEO needs a pre-training step for $D_{\beta}$ that is also modeled by a neural network with parameter $\bu$. We additionally denote by $D_{\bu^{(l)}}$ the $l$th iterate of $D_{\beta}$. With the same notation in Supplementary~\ref{appen:alg_gsp}, the numerical algorithm for GEO appears in Algorithm~\ref{alg:geo}. 
\begin{algorithm}[th]
\caption{Generalized Equalized Odds}\label{alg:geo}
\KwData{Let $\{(x_i,a_i,y_i)\}_{i=1}^n$ be the set of training data set and obtain $\{a_i'\}_{i=1}^n$ by permuting $\{a_i\}_{i=1}^n$. Denote the $t$th iterate of model parameters by $\bu^{(t)}$, $\bw^{(t)}$  and $\bv^{(t)}$ and $\alpha$ is a learning rate. Fix $L$, $T$, and $T'$; set $t=0$ and $l=0$; and initialize $\bu^{(0)}$, $\bw^{(0)}$ and $\bv^{(0)}$.}
\KwResult{$h_{\bv^{(T)}}$}
\While{$l \leq L$}{
    $l=l+1$\\ 
    
    $\{(x_{(i)},a_{(i)},y_{(i)},a_{(i)}')\}_{i=1}^{n_b} = \text{MC}({\bf D}_n)$\\
    
    $\hat{R}_{\beta}(D_{\bu^{(l)}})=\frac{1}{n_b} \sum_{i=1}^{n_b} \left( \log D_{\bu^{(l)}}(a_{(i)},y_{(i)})  + \log (1-D_{\bu^{(l)}}(a'_{(i)},y_{(i)})) \right)$\\
    
    $\bu^{(l+1)} = \bu^{(l)} + \alpha \dfrac{\partial}{\partial \bu^{(l)}}\hat{R}_{\beta}(D_{\bu^{(l)}})$\\ 

}
\While{$t \leq T$}{
    $t=t+1$\\ 
    
    $\{(x_{(i)},a_{(i)},y_{(i)},a_{(i)}')\}_{i=1}^{n_b} = \text{MC}({\bf D}_n)$\\

    $t' = 0$\\ 
    \While{$t' \leq T'$}{
    
    $t' = t' + 1$ \\ 

    {\small $\hat{R}_F(h_{\bv};D_{\bw^{(t)}})=\frac{1}{n_b} \sum_{i=1}^{n_b} \left( \log D_{\bw^{(t)}}(h_{\bv^{(t)}}(x_{(i)}),a_{(i)},y_{(i)}) + \dfrac{D_{\bu^{(L)}}(a_{(i)},y_{(i)})}{1-D_{\bu^{(L)}}(a_{(i)},y_{(i)})}\log (1-D_{\bw^{(t)}}(h_{\bv^{(t)}}(x_{(i)}),a'_{(i)},y_{(i)})) \right)$}\\
    
    $\bw^{(t+1)} = \bw^{(t)} + \alpha \dfrac{\partial}{\partial \bw^{(t)}}\hat{R}_F(h_{\bv^{(t)}};D_{\bw^{(t)}})$\\ 
    }

    $\hat{R}(h_{\bv^{(t)}})=\frac{1}{n_b}\sum_{i=1}^{n_b} L(y_{(i)}, h_{\bv^{(t)}}(x_{(i)}))$ \\

    $\bv^{(t+1)} = \bv^{(t)} - \alpha \dfrac{\partial}{\partial \bv^{(t)}} (\hat{R}(h_{\bv^{(t)}}) +  \lambda \hat{R}_F(h_{\bv^{(t)}};D_{\bw^{(t)}}))$ \\ 
}
\end{algorithm}

\subsection{Estimation performance from $D_{\beta}$ on a toy experiment}
\label{appen:geo_toy_exp}

We test the estimation performance of $D_{\beta}$ for the density ratio estimator $\beta$. Suppose $p(y|a)=P(Y=y|A=a)=(1+\exp(-a))^{-1}$ with $P(A=a)=0.5$ for $y,a\in\{0,1\}$. $D_{\beta}$ is assumed to have 2 dense hidden layers with 64 nodes and Relu activation functions, and its output layer has a single dimension with Sigmoid activation. In the total number of 10000 training iterations with 100 minibatch size, the last iterate is selected as the point estimate of $p(y|a)/p(y)$. Based on 5 replicated experiments, we calculate the averages and the standard deviation of the point estimates. As it is shown in Table~\ref{tab:weight}, $\hat{\beta}$ successfully estimates the true density ratios.

\begin{table}[th]
\centering
\caption{Estimation performance for $\hat{\beta}$ in the toy example. The point estimate is found by averaging 5 outputs. Std. implies the standard deviation calculated from the 5 outputs. }
\begin{tabular}{l|l|l|l|l}
\toprule
& ~~~~$p(1|1)/p(1)$ & ~~~~$p(1|0)/p(1)$ & ~~~~$p(0|1)/p(0)$ & ~~~~$p(0|0)/p(0)$ \\ 
\hline
\hline
True &  1.1877    &     0.8123     &   0.6995   &    1.3005  \\
Point Estimate (Std.)  &  1.1810 (0.0077)   &     0.8126   (0.0029)    &   0.7082  (0.0018)   &    1.2955  (0.0048)  \\
\bottomrule
\end{tabular}
\label{tab:weight}
\end{table}

\newpage

\section{DETAILS OF SECTION~\ref{sec:theory}}
\label{appen:theory}

\subsection{Proof of Theorem~\ref{thm:ee}}

Our theory specifically deals with binary cross-entropy and mean absolute error for binary and continuous outcomes respectively. The overall proof strategy is to check the bounded difference condition to use McDiarmid's inequality. 

The estimation error can be defined as  
\begin{align}
    |d(\hat{h}^*;\lambda) - \inf_{h \in {\cal H}} d(h;\lambda)|,
    \label{thm:esterror}
\end{align}
where $\hat{h}^*=\arg_h \min \hat{d}(h;\lambda)$. The estimation error is further decomposed as follows. 
\begin{align}
    d(\hat{h}^*;\lambda) - \inf_{h} d(h;\lambda)= d(\hat{h}^*;\lambda) -  d(h^*;\lambda)  &= d(\hat{h}^*;\lambda) - \hat{d}(\hat{h}^*;\lambda) \label{thm:bdd1} \\
    &\quad +\hat{d}(h^*;\lambda) -d(h^*;\lambda) \label{thm:bdd2} \\
    &\quad +\hat{d}(\hat{h}^*;\lambda) - \hat{d}(h^*;\lambda), \label{thm:bdd3}
\end{align}
where $h^*=\arg_h \min d(h;\lambda)$, and it is trivial to see $\eqref{thm:bdd3}\leq 0$.
The first line \eqref{thm:bdd1} is equivalent to
\begin{align*}
    d(\hat{h}^*;\lambda) - \hat{d}(\hat{h}^*;\lambda) = R(\hat{h}^*)-\hat{R}(\hat{h}^*) + \lambda(R_F(\hat{h}^*)-\hat{R}_F(\hat{h}^*)). 
\end{align*}
For \eqref{thm:bdd2}, we have 
\begin{align*}
   \hat{d}(h^*;\lambda)- d(h^*;\lambda) 
    = \hat{R}(h^*) - R(h^*) + \lambda(\hat{R}_F(h^*) - R_F(h^*)).
\end{align*}
Therefore, the estimation error is bounded by
\begin{align}
\label{appn:decomp}
    |d(\hat{h}^*;\lambda) - \inf_{h} d(h;\lambda)|\leq 2 \underbrace{\sup_{h}|R(h)-\hat{R}(h)|}_{\text{I}} + 2 \lambda \underbrace{\sup_{h}|R_F(h)-\hat{R}_F(h)|}_{\text{II}}.
\end{align}
Since $\hat{R}(h)$ is an empirical risk function, we denote $T((X_1,Y_1),\dots,(X_n,Y_n))=\sup_h |R(h)-\hat{R}(h)|$. For the binary cross-entropy loss function, i.e., $L(Y_i, h(X_i))= Y_i \log \sigma(h(X_i)) + (1-Y_i)\log (1-\sigma(h(X_i)))$, the bounded difference of the $i$th differing variable can be bounded by
\begin{align}
\label{appn:bce}
    |T((X_1,Y_1),\dots,(X_i,Y_i),\dots,(X_n,Y_n))&-T((X_1,Y_1),\dots,(X_i^{\dagger},Y_i^{\dagger}),\dots,(X_n,Y_n))| \notag \\
    &\leq \sup_{h}|\dfrac{1}{n}(L(Y_i,h(X_i))-L(Y_i^{\dagger},h(X_i^{\dagger})))|,\notag\\
    &\leq \dfrac{1}{n}\sup_{h} | Y_i \log \sigma (h(X_i)) + (1-Y_i)\log (1-\sigma (h(X_i))) - \notag\\ 
    &Y_i^{\dagger} \log \sigma (h(X_i^{\dagger})) - (1-Y_i^{\dagger})\log (1-\sigma (h(X_i^{\dagger}))) |. 
\end{align}
Without loss of generality, let's consider (i) $Y_i = Y_i^{\dagger}=1$, (ii) $Y_i=Y_i^{\dagger}=0$, and (iii) $Y_i=1$ and $Y_i^{\dagger}=0$. Since the sigmoid function $\sigma$ is a 1-Lipschitz function,  (i) upper bounds 
\begin{align*}
    |\log \sigma (h(X_i)) - \log \sigma (h(X_i^{\dagger})) | &\leq \dfrac{1}{\gamma_0} |\sigma (h(X_i)) - \sigma (h(X_i^{\dagger}))|, \\
    & \leq \dfrac{1}{\gamma_0}|h(X_i) - h(X_i^{\dagger})|, \\
    & \leq \dfrac{1}{\gamma_0}\prod_{j=1}^{g}M_{v}(j)\prod_{k=1}^{g-1}K_{\psi}(k)\times 2B. 
\end{align*}
The first inequality comes from the Lipschitz property of the logarithm whose domain is bounded below by a positive constant. The Cauchy-Schwarz inequality and the Lipschitz conditions of the activation functions lead the last inequality. Similarly, for (ii), we have 
\begin{align*}
    |\log (1- \sigma (h(X_i))) - \log (1-\sigma (h(X_i^{\dagger}))) | &\leq \dfrac{1}{1-\gamma_1} |\sigma (h(X_i)) - \sigma (h(X_i^{\dagger}))|, \\
    & \leq \dfrac{1}{1-\gamma_1}\prod_{j=1}^{g}M_{v}(j)\prod_{k=1}^{g-1}K_{\psi}(k)\times 2B, 
\end{align*}
For (iii), we obtain
    \begin{align*}
    |\log \sigma (h(X_i)) - \log (1-\sigma (h(X_i^{\dagger}))) | &\leq \dfrac{1}{\min \{\gamma_0, 1-\gamma_1\}} |\sigma (h(X_i)) + \sigma (h(X_i^{\dagger})) -1| \leq \dfrac{|2\gamma_1-1|}{\min\{\gamma_0, 1-\gamma_1\}}. 
\end{align*}
Therefore, the upper bound \eqref{appn:bce} is represented as 
\begin{align*}
    \max\left\{\dfrac{|2\gamma_1-1|}{n\gamma_{0,1}},\dfrac{1}{n\gamma_{0,1}}\prod_{j=1}^{g}M_{v}(j)\prod_{k=1}^{g-1}K_{\psi}(k)\times 2B\right\}, 
\end{align*}
where $\gamma_{0,1}=\min\{\gamma_0, 1-\gamma_1\}$. 

On the one hand, if the underlying loss function has the mean absolute error, i.e., $L(Y_i,h(X_i))=|Y_i-h(X_i)|$, the upper bound of the bounded difference is 
\begin{align*}
    |T((X_1,Y_1),\dots,(X_i,Y_i),\dots,(X_n,Y_n))&-T((X_1,Y_1),\dots,(X_i^{\dagger},Y_i^{\dagger}),\dots,(X_n,Y_n))|\\
    &\leq \sup_{h}|\dfrac{1}{n}(L(Y_i,h(X_i))-L(Y_i^{\dagger},h(X_i^{\dagger})))|,\\
    &\leq \dfrac{1}{n}\sup_{h}( |Y_i - Y_i^{\dagger}| + |h(X_i)-h(X_i^{\dagger})| ),\\
    &\leq \dfrac{1}{n} \left(1 + \prod_{j=1}^{g}M_{v}(j)\prod_{k=1}^{g-1}K_{\psi}(k)\times 2B \right). 
\end{align*}
Next, we take expectation to (I) in \eqref{appn:decomp} with respect to the random samples. We obtain 
\begin{align*}
    \expect_{X,Y} \sup_{h}|R(h)-\hat{R}(h)|&= \expect_{X,Y} \sup_{h}|\expect_{\tilde{X},\tilde{Y}}\dfrac{1}{n}\sum_{i=1}^n L(\tilde{Y}_i,h(\tilde{X}_i))-\dfrac{1}{n}\sum_{i=1}^n L(Y_i,h(X_i))|,\\
    &\leq \expect_{X,Y,\tilde{X},\tilde{Y}} \sup_{h}|\dfrac{1}{n}\sum_{i=1}^n L(\tilde{Y}_i,h(\tilde{X}_i))-\dfrac{1}{n}\sum_{i=1}^n L(Y_i,h(X_i))|,\\
    &=\expect_{X,Y,\tilde{X},\tilde{Y},
    \epsilon} \sup_{h}|\dfrac{1}{n}\sum_{i=1}^n \epsilon_i (L(\tilde{Y}_i,h(\tilde{X}_i))-L(Y_i,h(X_i)))|,\\
    &\leq 2\expect_{X,Y,\epsilon}\sup_{h}|\dfrac{1}{n}\sum_{i=1}^n\epsilon_i L(Y_i,h(X_i))|:=2{\cal R}({\cal L}),
\end{align*}
where $\epsilon_i \sim {\rm Unif}\{-1,1\}$ are i.i.d, $(\tilde{X}_i,\tilde{Y}_i)$ are i.i.d. copies (ghost samples) of $(X,Y)$, and ${\cal L}:=\{L(y,h_{\bv}(x)):\bv \in V\}$ is a function class of the given loss $L$. 

Therefore, McDiarmid's inequality implies 
\begin{align}
\label{thm:main_loss}
    \text{I}\leq 2{\cal R}({\cal L}) + F_{\bV,\psi,B,\gamma_0,\gamma_1}\sqrt{\dfrac{\log(1/\delta)}{2n}}, 
\end{align}
 with the 1-$\delta$ probability, where ${\cal R}$ is the Rademacher complexity of ${\cal L}$, and the involved constant is 
\begin{align*}
    F_{\bV,\psi,B,\gamma_0,\gamma_1} = \begin{cases}
                           \max\left\{\dfrac{|2\gamma_1-1|}{\gamma_{0,1}}, \dfrac{1}{\gamma_{0,1}} \prod_{j=1}^{g}M_{v}(j)\prod_{k=1}^{g-1}K_{\psi}(k)\times 2B\right\}
                            \quad \text{if } L \text{               is cross-enstropy}, \\
                            \left(1 + \prod_{j=1}^{g}M_{v}(j)\prod_{k=1}^{g-1}K_{\psi}(k)\times 2B \right)  
                            \quad \text{if } L \text{ is mean absolute difference}.
  \end{cases}
\end{align*}        

Similarly, we can upper bound (II).  First, we check the bounded difference condition that is induced by the neural penalty, i.e., 
\begin{align*}
\text{II} 
&\leq \underbrace{\sup_{h,D} [|\expect_{X,A}\log D(h(X),A)- \hat{\expect} \log D(h(X),A)|]}_{U_1((X_1,A_1),\dots,(X_i,A_i),\dots,(X_n,A_n))} \\
&+ \underbrace{\sup_{h,D} [|\expect_{X,A'}\log (1-D(h(X),A'))- \hat{\expect} \log (1-D(h(X),A'))|]}_{U_2((X_1,A'_1),\dots,(X_i,A'_i),\dots,(X_n,A'_n))}.
\end{align*}

Then, for any differing the $i$th coordinate,  
\begin{align*}
    |U_1((X_1,A_1),\dots,(X_i,A_i),\dots,(X_n,A_n))&-U_1((X_1,A_1),\dots,(X_i^{\dagger},A_i^{\dagger}),\dots,(X_n,A_n))| \\
    &\leq \dfrac{1}{n} \sup_{h,D} |\log D(h(X_i),A_i)- \log D(h(X_i^{\dagger}),A_i^{\dagger})|, \\
    & \leq \dfrac{1}{n\nu_0}|\sigma(f(h(X_i),A_i)) - \sigma(f(h(X_i^{\dagger}),A_i^{\dagger}))|,\\
    &\leq \dfrac{1}{n \nu_0 }\prod_{k=1}^{d}M_w(k)\prod_{j=1}^{d-1}K_{\kappa}(j)\times ||[h(X_i),A_i]-[h(X_i^{\dagger}),A_i^{\dagger}]||, \\
    &\leq \dfrac{1}{n\nu_0}\prod_{k=1}^{d}M_w(k)\prod_{j=1}^{d-1}K_{\kappa}(j)\times \sqrt{l+4B^2 \left( 
\prod_{k=1}^g M_v(k) \prod_{l=1}^{g-1}K_{\psi}(l) \right)^2}, \\ 
&:= \dfrac{F_{\bW,\bV,B,\kappa,\psi,l}}{n\nu_0}.
\end{align*}
Likewise, we get the similar result for $U_2$ as follows, 
\begin{align*}
    |U_2((X_1,A'_1),\dots,(X_i,A'_i),\dots,(X_n,A'_n))&-U_2((X_1,A'_1),\dots,(X_i^{\dagger},(A_i')^{\dagger}),\dots,(X_n,A'_n))| \\
    &\leq \dfrac{F_{\bW,\bV,B,\kappa,\psi,l}}{n(1-\nu_1)}.
\end{align*}

In addition, 
\begin{align*}
    \expect_{X,A} U_1((X_1,A_1),\dots,(X_i,A_i),\dots,(X_n,A_n)) 
    & \leq \expect_{X,A} \expect_{\tilde{X},\tilde{A}}\sup_{h,D}|\dfrac{1}{n}\sum_{i=1}^n \log D(h(\tilde{X}_i),\tilde{A}_i)-\log D(h(X_i),A_i)|, \\
    & \leq \dfrac{1}{\nu_0} \expect_{X,A} \expect_{\tilde{X},\tilde{A}}\sup_{h,f}|\dfrac{1}{n}\sum_{i=1}^n f(h(\tilde{X}_i),\tilde{A}_i)- f(h(X_i),A_i)|,\\
    & = \dfrac{1}{\nu_0} \expect_{X,A} \expect_{\tilde{X},\tilde{A},\epsilon}\sup_{h,f}|\dfrac{1}{n}\sum_{i=1}^n \epsilon_i (f(h(\tilde{X}_i),\tilde{A}_i)-f(h(X_i),A_i))|,\\
    &\leq \dfrac{2}{\nu_0}\expect_{X,A,\epsilon}\sup_{h,f}|\dfrac{1}{n}\sum_{i=1}^n\epsilon_i f(h(X_i),A_i)|:=\dfrac{2}{\nu_0} {\cal R}({\cal D}),
\end{align*}
where $\epsilon_i \sim {\rm Unif}\{-1,1\}$, are i.i.d., $(\tilde{X}_i,\tilde{A}_i)$ are i.i.d. copies (ghost sample) of $(X,A)$, and ${\cal D}=\{f_{\bw}(h_{\bv}(x),a):\bw \in \bW, \bv\in\bV \}$ is a compositional function class. Therefore, with the probability $1-\delta$,
\begin{align*}
    U_1((X_1,A_1),\dots,(X_i,A_i),\dots,(X_n,A_n))\leq \dfrac{2}{\nu_0} {\cal R}({\cal D})+ \dfrac{1}{\nu_0}F_{\bW,\bV,B,\kappa,\psi,l}\sqrt{\dfrac{\log(1/\delta)}{2n}}.
\end{align*}
Similarly for $U_2$, with $1-\delta$ probability, 
\begin{align*}
    U_2((X_1,A'_1),\dots,(X_i,A'_i),\dots,(X_n,A'_n))\leq \dfrac{2}{1-\nu_1} {\cal R}({\cal D})+ \dfrac{1}{(1-\nu_1)}F_{\bW,\bV,B,\kappa,\psi,l}\sqrt{\dfrac{\log(1/\delta)}{2n}}. 
\end{align*}

Therefore, with $1-2\delta$ probability, 
\begin{align*}
    \text{II}\leq 2\left(\dfrac{1}{\nu_0}+\dfrac{1}{1-\nu_1}\right){\cal R}({\cal D}) + \left(\dfrac{1+\nu_0-\nu_1}{\nu_0 (1-\nu_1)}\right) F_{\bW,\bV,B,\kappa,\psi,l}\sqrt{\dfrac{\log(1/\delta)}{2n}}.
\end{align*}
Consequently, by combining I and II, the estimation error is bounded above by 
\begin{align*}
|d(\hat{h}^*;\lambda) - &\inf_{h} d(h;\lambda)| \leq \\
    &4{\cal R}({\cal L}) + 2F_{\bV,\psi,B}\sqrt{\dfrac{\log(1/\delta)}{2n}} + 2\lambda \left(F_{\nu_0,\nu_1}{\cal R}({\cal D}) +
 F_{\bW,\bV,B,\kappa,\psi,l,\nu_0,\nu_1}\sqrt{\dfrac{\log(1/\delta)}{2n}}\right),
\end{align*}
with $1-3\delta$ probability, where $F_{\nu_0,\nu_1} = 2\left(\frac{1}{\nu_0}+\frac{1}{1-\nu_1}\right)$ and 
$F_{\bW,\bV,B,\kappa,\psi,l,\nu_0,\nu_1} = \left(\frac{1+\nu_0-\nu_1}{\nu_0 (1-\nu_1)}\right)\times F_{\bW,\bV,B,\kappa,\psi,l}$.
    
\subsection{Proof of Corollary \ref{thm:cor}}

Let $\hat{h}^* = \arg\min \hat{d}(h;\lambda)$, $h^*_0 = \arg\min d(h;\lambda=0)$, and $h^* = \arg\min d(h;\lambda)$. Let's denote by $\Delta(h_0^*, h^*):=R_F(h_0^*) - R_F(h^*)$ the amount of decision discrimination in the population with respect to the sensitive information. Since $d(h^*;\lambda)\leq d(h_0^*;\lambda)$, we obtain 
\begin{align*}
d(\hat{h}^*;\lambda=0)-\inf_h d(h;\lambda=0)&=R(\hat{h}^*)-R(h_{0}^*),\\
              &=R(\hat{h}^*)-R(h^*)+R(h^*)-R(h_{0}^*), \\
              &\leq R(\hat{h}^*)-R(h^*) + \lambda \Delta(h_0^*, h^*).
\end{align*}
Similar to $\eqref{thm:bdd1}\sim \eqref{thm:bdd3}$, the loss of utility has the following upper bound 
\begin{align*}
|d(\hat{h}^*;\lambda=0)-\inf_h d(h;\lambda=0)|&\leq 2\underbrace{\sup_h |R(h)-\hat{R}(h)|}_{\text{I}}+\lambda \Delta(h_0^*, h^*),
\end{align*}
where $\text{I}$ has the same upper bound \eqref{thm:main_loss}. 

\newpage

\section{DETAILS OF SIMULATION}

This section explains simulation setups and delivers additional empirical studies to justify the performance of our method. The outline of this section is as follows. 
\begin{itemize}
    \item \ref{appen:sim_setting} explains the simulation settings concretely.  
    \item \ref{appen:simul_sp} displays all other simulation results omitted in the main text such as simulation results for (generalized) statistical parity in Scenario II and (generalized) equalized odds in Section I, trade-off figures, and additional tables summarizing fairness scores.
    \item \ref{appen:simul_scen3} compares our model to HGR and CON in Scenario III. 
    \item \ref{sec:robust_beta} discusses the estimation stability of our method when the density ratio estimator (for separation) is poor. 
    \item \ref{sec:fair_rep} illustrates our method outperforms in having fair representation. 
\end{itemize}

\label{appen:sim}
\subsection{Overall simulation setting}
\label{appen:sim_setting}
\paragraph{Target objective function} We first clarify that the notation $(1-\lambda)L_M+\lambda L_F$ in the manuscript. For the competing methods, we refer to the notation of their original papers. All methods below share $\hat{R}(h)$ to express an empirical risk function. 
\begin{itemize}
    \item For Ours, 
    \begin{align*}
        L_M = \hat{R}(h), \quad L_F = \hat{R}_F(h;D).
    \end{align*}
    \item CON has the same $\hat{R}(h)$ and $\hat{R}_F(h;D)$ but it uses $A'\sim P(A|Y)$.  
    \item For HGR \citep{lee:etal:22}, 
    \begin{align*}
        L_M &= \hat{R}(h), \\
        L_F &= {\text{HGR}_{\text{soft}}}(h(X),A\otimes Y)- {\text{HGR}_{\text{soft}}}(h(X), Y)\quad \text{for EO},\\
        L_F &= {\text{HGR}_{\text{soft}}}(h(X),A) \quad \text{for SP}.
    \end{align*}
    Refer to the original paper to see the exact form of ${\text{HGR}_{\text{soft}}}$. 
    \item For KDE \citep{cho:etal:20}, 
    \begin{align*}
        L_M &= \hat{R}(h), \\
        L_F &= {\text{DDP}} \quad \text{for SP}, \\
        L_F &= {\text{DEO}} \quad \text{for EO}, 
    \end{align*}
     Refer to the original paper to see the exact expression of DDP and DEO.
    \item For NEU \citep{du:etal:21}, 
    \begin{align*}
        L_M = {\cal L}_{\text{MSE}}, \quad L_F = {\cal L}_{\text{Smooth}}.
    \end{align*}
    Refer to the original paper to see the exact expression of ${\cal L}_{\text{MSE}}$ and ${\cal L}_{\text{Smooth}}$. It is assumed that the sensitive information is available during the training phase. For NEU, the model $h$ is pre-trained by minimizing $\hat{R}(h)$ with an early-stopping procedure. Then the discriminatory $h$ is trained to minimize $(1-\lambda)L_M+\lambda L_F$ to make it debiased. 
\end{itemize}

\paragraph{Network architectures for the simulation studies} All methods have the same neural network model for $h$. More specifically,  
\begin{enumerate}
    \item[A.] {\bf Scenario I \& II (binary classification with cross-entropy)} 
        \begin{itemize}
            \item $h$ has [Dense(64)-BN-ReLU]*3-[Dense(1)-Sigmoid].
            \item $D$ has [Dense(64)-BN-ReLU]*2-[Dense(1)-Sigmoid]. 
            \item $D_{\beta}$ has [Dense(64)-ReLU]*2-[Dense(1)-Sigmoid] \quad (\text{for separation}). 
        \end{itemize}
        For $\text{HGR}_{\text{soft}}$, two neural networks are needed, and each of them is set to  [Dense(64)-BN-ReLU]*2-[Dense(1)]. As a result, HGR employs four auxiliary networks in total for separation and two networks for independence. For NEU, $h$'s first output layer is neutralized. For CON in Scenario II, the discriminator has the same structure with $D_{\beta}$ while the generator $G=[G_D,G_C]$ with $G_D=G_B$-$\text{Dense(1)-Sigmoid}$ and $G_C=G_B$-$\text{Dense(1)}$ where $G_B=$[Concat(Unif(2,[0,1]),$Y$)-Dense(64)-ReLU-Dense(64)-ReLU]. Unif(2,[0.1]) denotes a 2-dimensional uniform random variable each of which is in $[0,1]$. The binary value is determined by binning the outputs of $G_D$ with the threshold 0.5. 
        
    \item[B.] {\bf Scenario III (regression with mean absolute error)} 
        \begin{itemize}
            \item $h$ has [Dense(16)-BN-ReLU]*3-[Dense(1)].
            \item $D$ has [Dense(16)-BN-ReLU]*2-[Dense(1)-Sigmoid].
            \item $D_{\beta}$ has [Dense(16)-ReLU]*2-[Dense(1)-Sigmoid] \quad (\text{for separation}).
        \end{itemize}
        $\text{HGR}_{\text{soft}}$ has two [Dense(16)-BN-ReLU]*2-[Dense(1)]. For CON, the generator only has a continuous output $G=$[Concat(Unif(2,[0,1]),$Y$)-Dense(16)-ReLU-Dense(16)-Dense(1)]. 
\end{enumerate}

\paragraph{Hyperparameter} All models experimented 5 times for each $\lambda \in \{0.1,0.3,0.5,0.7,0.9\}$ based on 80\% training and 20\% validation set. Seed numbers are specified such that all models in the comparison are trained on the same data sets for each $\lambda$. For optimization, we adopt stochastic gradient descent with a learning rate of 0.005. For SBP and HGR, the number of training iterations for the maximization part of both algorithms, e.g., the notation $T'$ in our work (Algorithm~1 and 2), is set to 1. The evaluation metrics are calculated every 100 iterations. Each metric is basically expressed as a sum, but it is equivalent to finding an average value. To see other details including a mini-batch size or the number of epochs for all data sets, please refer to the shared code scripts.  

\paragraph{Training time comparison shown in Table~\ref{tab:tr_time}} NEU, SBP, and CON require pre-training courses, i.e., pre-training of $h$ for NEU, of $\hat{\beta}$ for SBP, the GAN model for CON.  As a possible case of NEU, pre-training of $h$ may need substantial training time or longer evaluation epochs for using early stopping. In the case of CON, learning the generator gets longer as the complexity of $A|Y$ increases. Tables~\ref{tab:tr_time} and \ref{tab:tr_time2} present the training times including/excluding the pre-training courses during early iterations for EO and SP to the Adult data. The same pattern is observed in other data sets, so they are omitted. All experiments, written in Tensorflow 2.4.0, run on CentOS 7 featuring Nvidia A30 GPU and 192GB of RAM.

\begin{table}[hbt!]
\centering
\caption{Training times (mins) for the first 1000 iterations on A30 GPU for independence in \textbf{Adult}. Note NEU needs the pre-training course while others do not, and KDE cannot be applied to Scenario II.}
\begin{tabular}{c|cc|c}
\toprule
 & \multicolumn{2}{c|}{Without pre-training}& \multicolumn{1}{c}{With pre-training} \\ 
 \hline
\multicolumn{1}{c|}{Model} &  Scenario I  &  Scenario II    &     Scenario II \\ 
 \hline
 \hline
\multicolumn{1}{c|}{SBP} &  0.23 (0.02)   & 0.56 (0.05)   & -  \\
\multicolumn{1}{c|}{HGR} &  0.24 (0.02)   & 0.59 (0.06)   & -  \\
\multicolumn{1}{c|}{KDE} &  0.88 (0.07)   &  -            & -  \\
\multicolumn{1}{c|}{NEU} &  1.01 (0.09)   & 1.46 (0.14)   & 3.61 (0.73)  \\
\bottomrule
\end{tabular}
\label{tab:tr_time2}
\vspace{-0.3cm}
\end{table}

\subsection{More simulation results in Scenario I \& II}
\label{appen:simul_sp}

\begin{figure}[hbt!]
    \centering
    \includegraphics[trim={2.5cm 0 0 0},clip,scale=0.40]{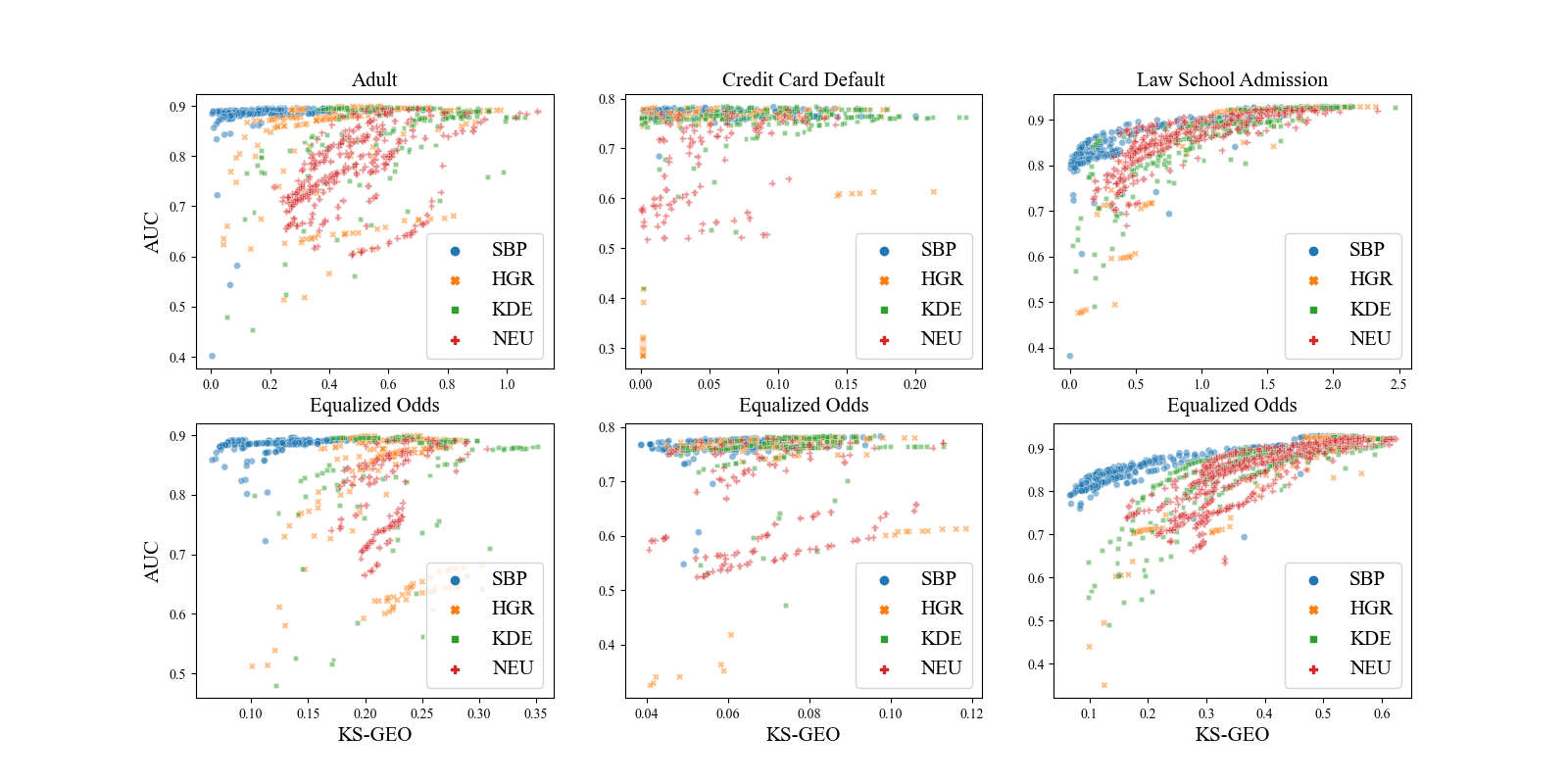}
    \includegraphics[trim={2.5cm 0 0 0},clip,scale=0.40]{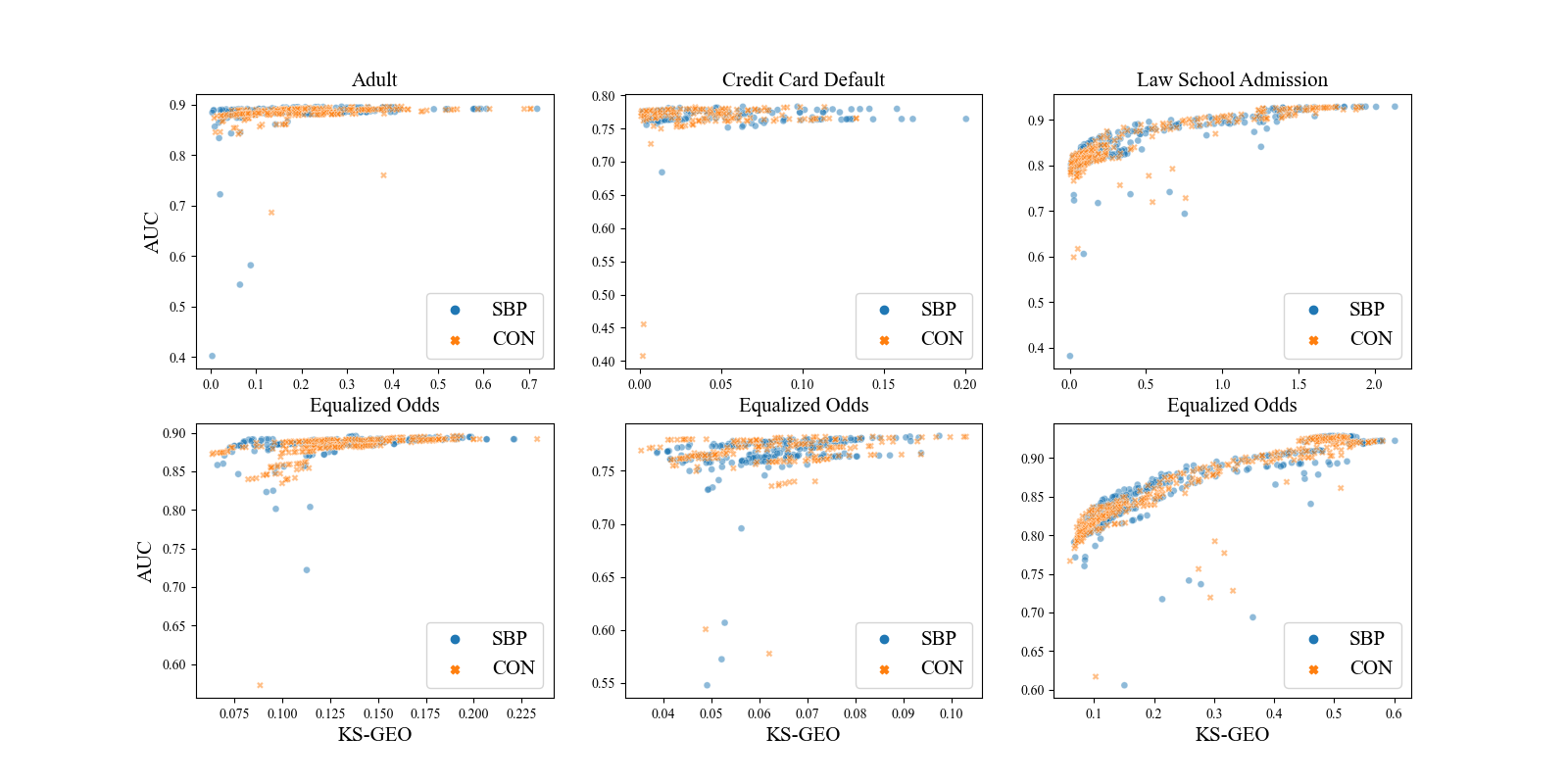}
    \caption{(Scenario I) Pareto frontiers: the first row includes pairs of EO and AUC, and the second row shows pairs of KS-GEO and AUC from 5 experiments for each $\lambda$. Since SBP and CON are similar, they are directly compared in the below figure. SBP and CON illustrate better results than others in Adult and Law School Admission as they are tightly in the upper-left corner.}
    \label{sim:s1eo}
\end{figure}

\begin{figure}[hbt!]
    \centering
    \includegraphics[trim={2.5cm 0 0 0},clip,scale=0.46]{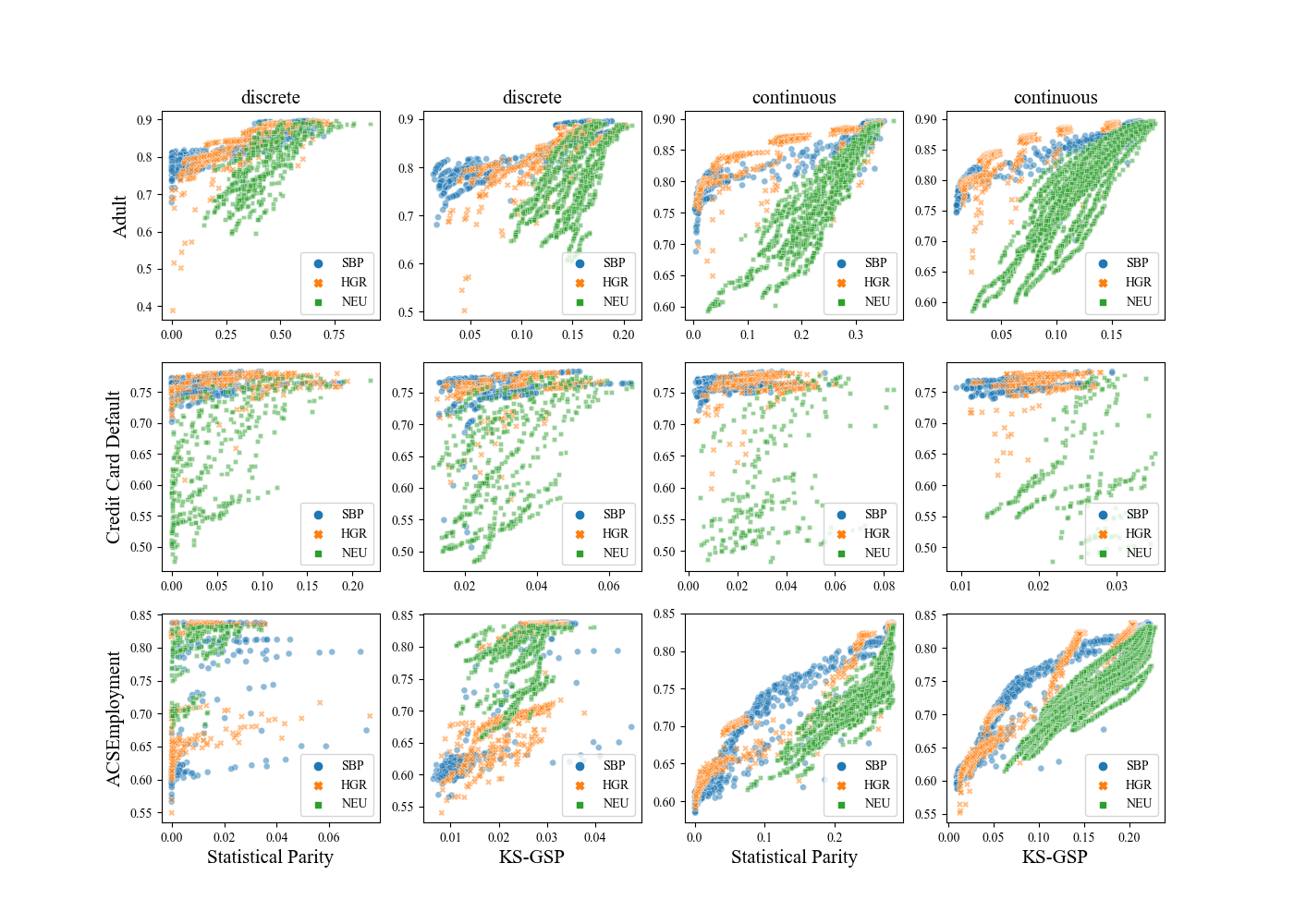}
    \caption{(Scenario II) Pareto frontiers: the first and the second row correspond to Adult, Credit Card Default, and ACSEmployment respectively from 5 experiments for each $\lambda$. SBP is superior to both NEU and HGR overall but comparable to HGR in Adult. Note CON cannot handle statistical parity.}
    \label{sim:s2sp}
\end{figure}

\begin{figure}[hbt!]
    \centering
    \includegraphics[trim={2.5cm 0 0 0},clip,scale=0.47]{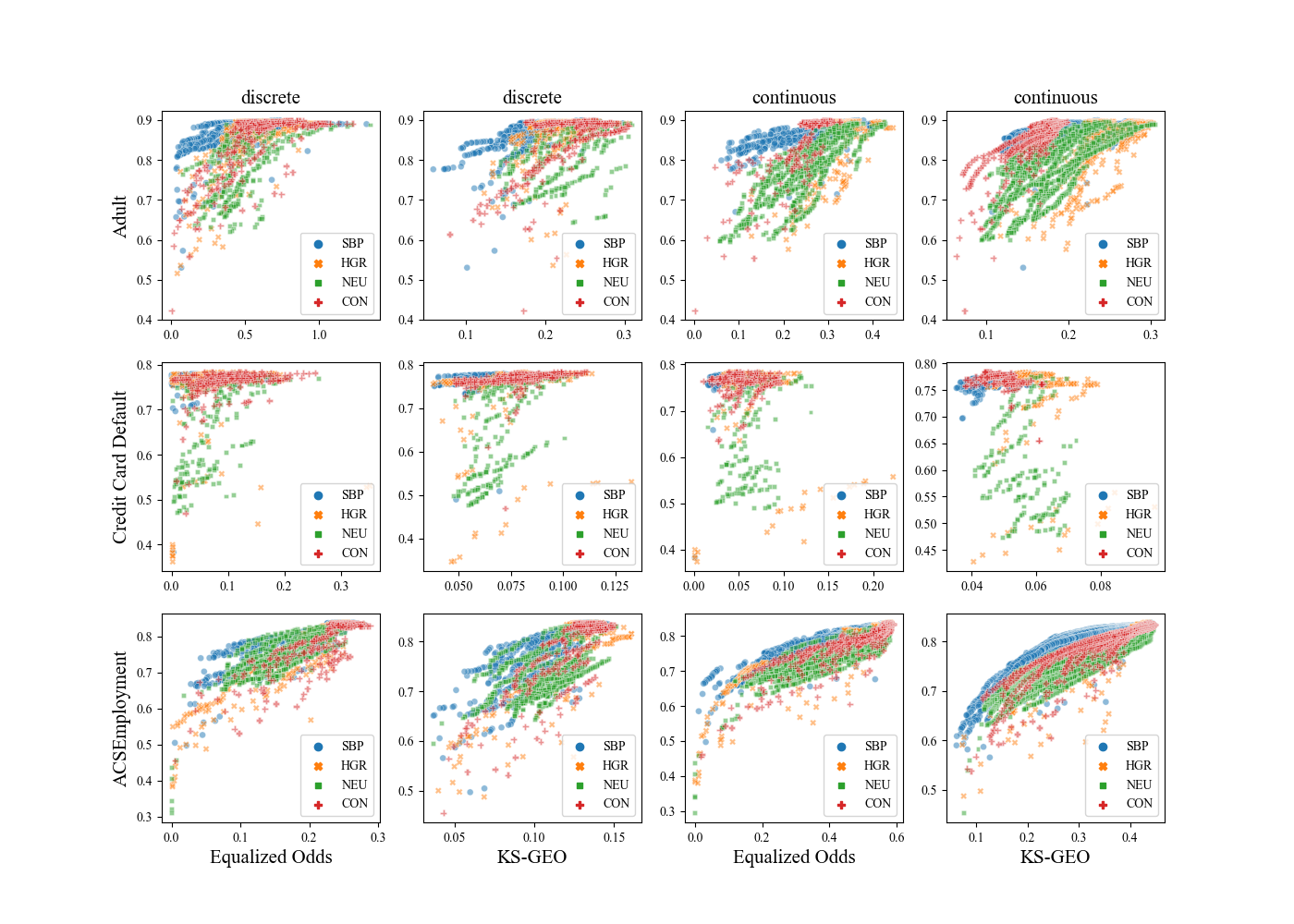}
    \caption{(Scenario II) Pareto frontiers: the first and the second row correspond to Adult, Credit Card Default, and ACSEmployment respectively from 5 experiments for each $\lambda$. Remarkably, SBP (ours) outperforms the competitors for the most part. Note KDE cannot handle continuous attributes.}
    \label{sim:s2eo}
\end{figure}

For the continuous variable, we calculate
\begin{align*}
    \text{SP} ~&=~ |{\cal A}^*|^{-1}\sum_{a\in {\cal A}^*}\left |\expect (\hat{Y}=1|A\leq a)/\expect (\hat{Y}=1) -1 \right|, \\
    \text{KS-GSP} ~&=~ |{\cal A}^*|^{-1}\sum_{a\in{\cal A}^*}\max_{h_x} |\hat{P}(\hat{h}^*(X)\leq h_x | A\leq a)-\hat{P}(\hat{h}^*(X)\leq h_x)|, 
\end{align*}
where ${\cal A}^*=\{\tilde{q}_{10},\dots,\tilde{q}_{90}\}$ with the $\tilde{q}_r:=\frac{r}{100}$th quantile of $A$. For EO and KS-GEO,
\begin{align*}
    \text{EO}&=|{\cal A}^*|^{-1}\sum_{y\in {\cal Y},a \in {\cal A}^*}\left| \frac{\expect (\hat{Y}=1|A\leq a,Y=y)}{\expect (\hat{Y}=1|Y=y)}-1\right|,\\
    \text{KS-GEO}&=|{\cal A}^*|^{-1}\sum_{y\in{\cal Y},a\in{\cal A}^*} \max_{h_x} |\hat{P}(\hat{h}^*(X)\leq h_x | A\leq a, Y=y)-\hat{P}(\hat{h}^*(X)\leq h_x |Y=y)|.
\end{align*}
Following the same manner in the manuscript, we collect the Pareto frontiers for all 5 experiments for each $\lambda$. It is possible that each experiment produces a set of Pareto solutions. Figures~\ref{sim:s1eo}, \ref{sim:s2sp}, and \ref{sim:s2eo} overlay those Pareto frontiers for all competitors and fairness measures. In Figure~\ref{sim:s1eo}, we observe that CON works as well as SBP in Scenario I. This is because $P(A|Y)$ can be easily estimated by the maximum likelihood method. 

\subsubsection{Trade-off curves by differing $\lambda$}

\begin{figure}[hbt!]
    \centering
    \includegraphics[trim={2.5cm 0 0 0},clip,scale=0.40]{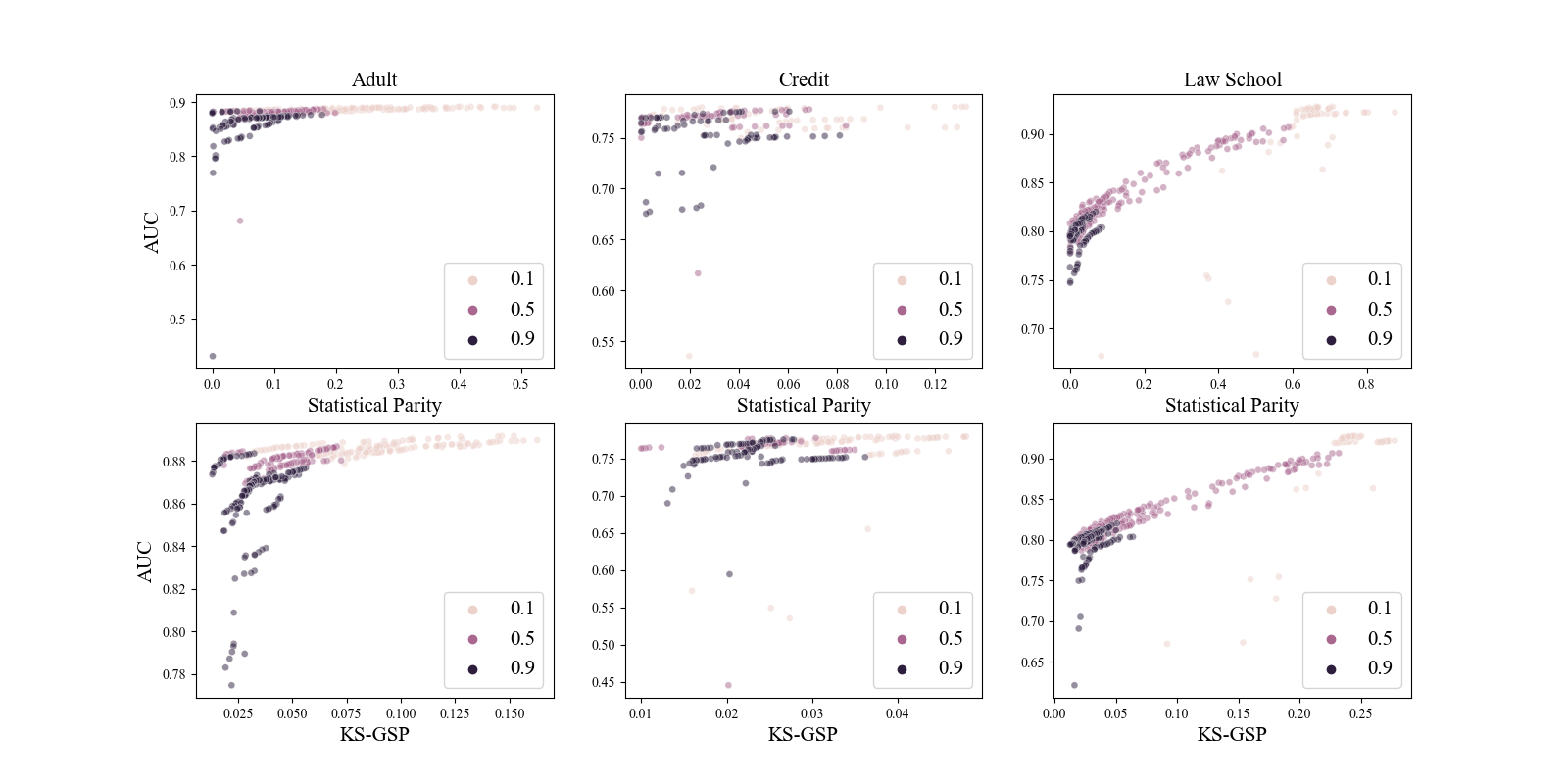}
    \caption{(Scenario I of SBP) Pareto frontiers of SP (and KS-GSP) and AUC by differing $\lambda$ from 0.1 to 0.9 for all 5 experiments.}
    \label{sim:s1todp}
\end{figure}

\begin{figure}[hbt!]
    \centering
    \includegraphics[trim={2.8cm 0 0 0},clip,scale=0.40]{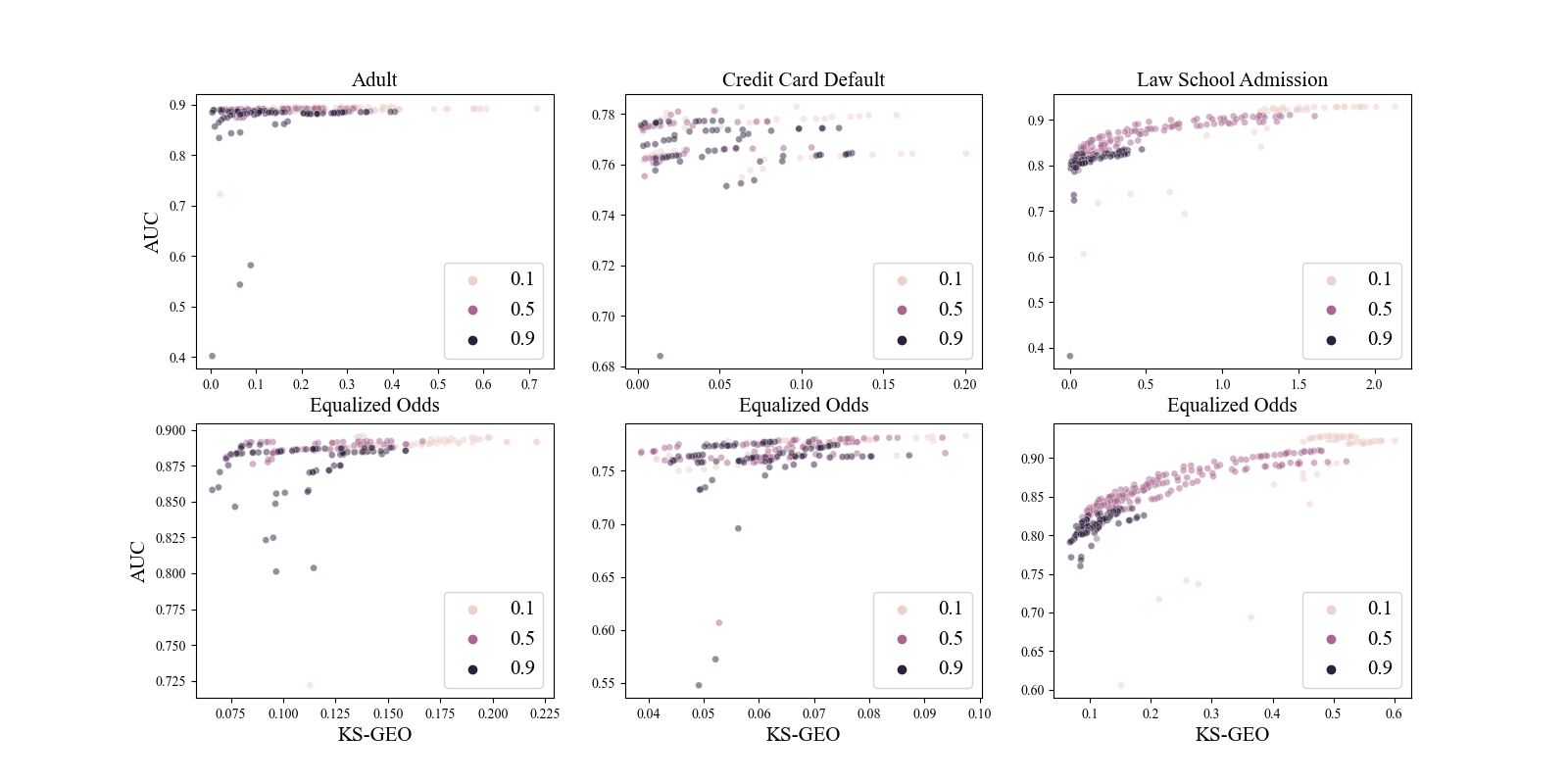}
    \caption{(Scenario I of SBP) Pareto frontiers of EO (and KS-GEO) and AUC by differing $\lambda$ from 0.1 to 0.9 for all 5 experiments.}
    \label{sim:s1toeo}
\end{figure}

Figure~\ref{sim:s1todp} shows that our method can capture the trade-off between utility and fairness in independence as $\lambda$ differs. For clear visualization, $\lambda$ is specifically chosen as 0.1, 0.5, and 0.9, respectively. The trade-off is also shown in separation (Figure~\ref{sim:s1toeo}).  

\subsubsection{Evaluation tables with other AUC thresholds}
\label{appen:tabs}
\begin{itemize}
    \item (Scenario I-SP) Table~\ref{tab:sce1_sp2} uses other thresholds compared to Table~\ref{tab:sce1_sp1}.
    \item (Scenario I-EO) Tables~\ref{tab:sce1_eo} and \ref{tab:sce1_eo2} show the tables with different thresholds. 
    \item (Scenario II-SP) Table~\ref{tab:s2sp2} show the results in SP. 
    \item (Scenario II-EO) Table~\ref{tab:s2eo2} uses other thresholds compared to Table~\ref{tab:s2eo}.
\end{itemize}

These extra tables show consistent results with the tables in the main text and with their corresponding figures. In general, our method outperforms the competing methods. The thresholds are chosen such that there are a sufficient number of fairness scores to calculate their averages and standard deviations from the 5 smallest scores. For instance, if the larger threshold is chosen such as AUC $\geq 0.9$ in Adult, there are no available fairness scores in all comparison methods.

\begin{table}[hbt!]
\caption{(Scenario I) Averages of the 5 smallest of SP/KS-GSPs whose AUCs are greater than the thresholds. Those scores are selected in the Pareto solutions appearing in Figure~\ref{sim:s1sp}. Standard deviations are in the parentheses next to the averages. Note CON is not available for SP.}
\label{tab:sce1_sp2}
\centering
\begin{tabular}{c|cc|cc|cc}
\toprule
& \multicolumn{2}{c|}{\textbf{Adult} (AUC $\geq$ 0.80)} & \multicolumn{2}{|c|}{\textbf{Cred. Card.} (AUC $\geq$ 0.70)} & \multicolumn{2}{|c}{\textbf{Law School.} (AUC $\geq$ 0.85)} \\
\hline
Model & SP   ($\downarrow$)    & KS-GSP  ($\downarrow$)   & SP   ($\downarrow$)        & KS-GSP     ($\downarrow$)     & SP    ($\downarrow$)         & KS-GSP  ($\downarrow$)      \\
\hline
\hline
SBP  & 0.001 ($\approx 0$) & 0.014 ($\approx 0$) &  $\approx 0$ ($\approx 0$)  & 0.011 (0.001) & 0.199 (0.032) &  0.112 (0.011) \\
HGR  & 0.070 (0.005)       & 0.040 ($\approx 0$) &  0.001 (0.001)              & 0.020 (0.001) & 0.123 (0.005) &  0.089 (0.001) \\
KDE  & 0.280 (0.018)       & 0.110 (0.005)       &  0.005 (0.003)              & 0.023 (0.001) & 0.252 (0.006) &  0.151 (0.001) \\
NEU  & 0.372 (0.018)       & 0.137 (0.002)       &  0.002 (0.001)              & 0.013 (0.001) & 0.385 (0.008) &  0.159 (0.006) \\ 
\bottomrule   
\end{tabular}
\end{table}

\begin{table}[hbt!]
\caption{(Scenario I) Averages of the 5 smallest of EO/KS-GEOs whose AUCs are greater than the thresholds. Those scores are selected in the Pareto solutions appearing in Figure~\ref{sim:s1eo}. Standard deviations are in the parentheses next to the averages.}
\label{tab:sce1_eo}
\centering
\begin{tabular}{c|cc|cc|cc}
\toprule
& \multicolumn{2}{c|}{\textbf{Adult} (AUC $\geq$ 0.85)} & \multicolumn{2}{|c|}{\textbf{Cred. Card.} (AUC $\geq$ 0.75)} & \multicolumn{2}{|c}{\textbf{Law School.} (AUC $\geq$ 0.85)} \\
\hline
Model & EO   ($\downarrow$)    & KS-GEO  ($\downarrow$)   & EO   ($\downarrow$)        & KS-GEO     ($\downarrow$)     & EO    ($\downarrow$)         & KS-GEO  ($\downarrow$)      \\
\hline
\hline
SBP  &  0.009 (0.005)    &   0.070 (0.002)   &  0.003 (0.001)      &  0.040 (0.001)      &    0.166 (0.027)     &   0.140 (0.005)      \\
CON  &  0.019 (0.008)   &    0.066 (0.002)  &   0.001 (0.001)      &  0.038 (0.002)      &    0.217 (0.033)     &   0.190 (0.009)       \\
HGR  &  0.164 (0.031)    &   0.170 (0.006)  &  0.002 (0.001)       &  0.047 (0.002)      &    0.891 (0.080)     &   0.409 (0.014)          \\
KDE  &  0.383 (0.013)    &   0.172 (0.003)    &  0.002 (0.001)     &  0.050 (0.001)      &    0.311 (0.128)     &   0.262 (0.006)          \\
NEU  &   0.450 (0.028)    &  0.205 (0.005)     &   0.030 (0.019)   &  0.046 (0.001)      &    0.421 (0.022)     &   0.299 (0.003)     \\ 
\bottomrule   
\end{tabular}
\end{table}

\begin{table}[hbt!]
\caption{(Scenario I) Averages of the 5 smallest of EO/KS-GEOs whose AUCs are greater than the thresholds. Those scores are selected in the Pareto solutions appearing in Figure~\ref{sim:s1eo}. Standard deviations are in the parentheses next to the averages.}
\label{tab:sce1_eo2}
\centering
\begin{tabular}{c|cc|cc|cc}
\toprule
& \multicolumn{2}{c|}{\textbf{Adult} (AUC $\geq$ 0.80)} & \multicolumn{2}{|c|}{\textbf{Cred. Card.} (AUC $\geq$ 0.70)} & \multicolumn{2}{|c}{\textbf{Law School.} (AUC $\geq$ 0.80)} \\
\hline
 Model & EO   ($\downarrow$)    & KS-GEO  ($\downarrow$)   & EO   ($\downarrow$)        & KS-GEO     ($\downarrow$)     & EO    ($\downarrow$)         & KS-GEO  ($\downarrow$)      \\
\hline
\hline
SBP  &  0.009 (0.005)  & 0.070 (0.003)  & 0.003 (0.001)        &  0.040 (0.001)   &  0.020 (0.010)      &  0.078 (0.001)     \\
CON  &  0.016 (0.007)  & 0.066 (0.002)  & 0.001 (0.001)        &  0.038 (0.002)   &  0.014 (0.003)      &  0.074 (0.002)    \\
HGR  &  0.127 (0.025)  & 0.167 (0.007)  & 0.002 ($\approx 0$)  &  0.047 (0.002)   &  0.587 (0.152)      &  0.365 (0.027)     \\
KDE  &  0.282 (0.064)  & 0.155 (0.009)  & 0.001 (0.001)        &  0.049 (0.001)   &  0.194 (0.016)      &  0.216 (0.007)     \\
NEU  &  0.364 (0.013)  & 0.182 (0.003)  & 0.016 (0.003)        &  0.046 (0.001)   &  0.380 (0.033)      &  0.245 (0.012)     \\ 
\bottomrule   
\end{tabular}
\end{table}

\begin{table}[hbt!]
\centering
\caption{(Scenario II) Averages of the 5 smallest of SP/KS-GSPs whose AUCs are greater than the thresholds. Those scores are selected by referring to Figure~\ref{sim:s2sp}. Standard deviations are in the parentheses next to the averages.}
\label{tab:s2sp2}
\begin{tabular}{c|c|cc|cc|cc}
\toprule
 && \multicolumn{2}{c|}{\textbf{Adult} (AUC $\geq$ 0.80)} & \multicolumn{2}{|c|}{\textbf{Cred. Card.} (AUC $\geq$ 0.75)} & \multicolumn{2}{|c}{\textbf{ACSEmpl.} (AUC $\geq$ 0.75)} \\
\hline
 Met. & Mod. &   Race  ($\downarrow$)     &  Age   ($\downarrow$)      &    Gender   ($\downarrow$)    &      Age  ($\downarrow$)    &  Gender   ($\downarrow$)  &   Age   ($\downarrow$)          \\
\hline
\hline
\multirow{3}{*}{SP} & SBP  & 0.002 (0.001) & 0.018 (0.002) &  $\approx 0$ ($\approx 0$)      & 0.004 (0.001) & 0.001 (0.001)    & 0.116 (0.002)      \\
& HGR  & 0.148 (0.006) & 0.019 (0.001) &  $\approx 0$ ($\approx 0$)      & 0.010 (0.001) & $\approx 0$ ($\approx 0$)    & 0.196 (0.004)     \\
& NEU  & 0.338 (0.018) & 0.216 (0.010) &  0.050 (0.011)                  & 0.040 (0.004) & $\approx 0$ ($\approx 0$)    & 0.216 (0.008)     \\ 
\hline
\hline
\multirow{3}{*}{\begin{tabular}[c]{@{}l@{}}KS-\\GSP\end{tabular}} & SBP  & 0.019 (0.002) & 0.015 ($\approx 0$) & 0.014 ($\approx 0$) & 0.010 ($\approx 0$)  &  0.019 (0.003)  &  0.078 (0.002)  \\
& HGR  & 0.062 (0.005) & 0.021 (0.001)       & 0.015 ($\approx 0$) & 0.012 ($\approx 0$)  &  0.017 (0.002) &  0.118 (0.001)  \\
& NEU  & 0.127 (0.001) & 0.088 (0.003)       & 0.025 (0.003)       & 0.026 ($\approx 0$)  &  0.012 ($\approx 0$)  &  0.140 (0.001)  \\ 
\bottomrule
\end{tabular}
\end{table}

\begin{table}[hbt!]
\centering
\caption{(Scenario II) Averages of the 5 smallest of EO/KS-GEOs whose AUCs are greater than the thresholds. Those scores are selected by referring to Figure~\ref{sim:s2eo}. Standard deviations are in the parentheses next to the averages.}
\label{tab:s2eo2}
\begin{tabular}{c|c|cc|cc|cc}
\toprule
 &  & \multicolumn{2}{c|}{\textbf{Adult} (AUC $\geq$ 0.85)} & \multicolumn{2}{|c|}{\textbf{Cred. Card.} (AUC $\geq$ 0.70)} & \multicolumn{2}{|c}{\textbf{ACSEmpl.} (AUC $\geq$ 0.80)} \\
\hline
 Met. & Mod. &   Race  ($\downarrow$)     &  Age   ($\downarrow$)      &    Gender   ($\downarrow$)    &      Age  ($\downarrow$)    &  Gender   ($\downarrow$)  &   Age   ($\downarrow$)          \\
\hline
\hline
\multirow{4}{*}{EO} & SBP  & 0.143 (0.013)  &  0.122 (0.002)  &  0.001 ($\approx 0$) & 0.016 (0.001)   &  0.168 (0.004)    &  0.373 (0.006)      \\
& CON  & 0.425 (0.017)  &  0.241 (0.002)  &  0.006 (0.003)       & 0.016 (0.003)   &  0.224 (0.004)    &  0.458 (0.031)     \\
& HGR  & 0.355 (0.010)  &  0.307 (0.002)  &  0.002 (0.001)       & 0.033 (0.001)   &  0.195 (0.004)    &  0.447 (0.002)      \\
& NEU  & 0.385 (0.022)  &  0.268 (0.012)  &  0.015 (0.010)       & 0.038 (0.004)   &  0.143 (0.008)    &  0.443 (0.017)      \\ 
\hline
\hline
\multirow{4}{*}{\begin{tabular}[c]{@{}l@{}}KS-\\GEO\end{tabular}} & SBP  & 0.147 (0.002)  &  0.124 (0.002) &  0.040 (0.001)  &  0.036 (0.001)    &  0.094  (0.001)   &  0.247 (0.002)    \\
& CON  & 0.177 (0.001)  &  0.138 (0.002) &  0.049 (0.001)  & 0.038 ($\approx 0$)&  0.115 (0.003)    &   0.312 (0.002)     \\
& HGR  & 0.162 (0.004)  &  0.221 (0.001) &  0.041 (0.003)  &  0.047 (0.001)    &  0.122  (0.001)   &  0.356 (0.005)    \\
& NEU  & 0.196 (0.008)  &  0.166 (0.004) &  0.053 (0.003)  &  0.051 (0.001)    &  0.101  (0.006)   &  0.334 (0.009)    \\ 
\bottomrule
\end{tabular}
\end{table}

\subsection{Simulation results (for both independence and separation) in Scenario III}
\label{appen:simul_scen3}
We further compare SBP with HGR and CON in Scenario III using {\bf Community and Crime}\footref{data:uci} data. Following \cite{lee:etal:22}, the number of violent crimes per population and the ratio of black people in the population are used as a continuous outcome and a sensitive variable respectively. For the evaluation of fairness, we consider the following metrics, 
\begin{align*}
    \text{SP} &= |{\cal A}^*|^{-1}\sum_{a\in {\cal A}^*}\left|\expect (\hat{h}^*(X)|A\leq a)/\expect (\hat{h}^*(X))-1 \right|, \\ 
    \text{EO} &= {|{\cal Y}^*|}^{-1}\sum_{y\in{\cal Y}^*}{|{\cal A}^*|}^{-1} 
    \sum_{a\in {\cal A}^*} \left|\expect (\hat{h}^*(X)|A\leq a,Y \leq y)/\expect (\hat{h}^*(X)|Y\leq y)-1\right|,\\
    \text{KS-GSP} &= |{\cal A}^*|^{-1} \sum_{a\in{\cal A}^*} \max_{h_x} |\hat{P}(\hat{h}^*(X)\leq h_x | A\leq a)-\hat{P}(\hat{h}^*(X)\leq h_x)|,\\
    \text{KS-GEO} &= {|{\cal Y}^*|}^{-1}\sum_{y\in{\cal Y}^*}{|{\cal A}^*|}^{-1} 
    \sum_{a\in {\cal A}^*}\max_{h_x} |\hat{P}(\hat{h}^*(X)\leq h_x | A\leq a, Y\leq y)-\hat{P}(\hat{h}^*(X)\leq h_x |Y\leq y)|,
\end{align*}
where ${\cal A}^*$ and ${\cal Y}^*$ are the sets of quantile values of $A$ and $Y$; ${\cal Y}^*=\{\tilde{b}_{10},\dots,\tilde{b}_{90}\}$ with the $\tilde{b}_r:=\frac{r}{100}$th quantile of $Y$.

\begin{figure}[h]
    \centering
    \includegraphics[trim={2.5cm 0 0 0},clip,scale=0.47]{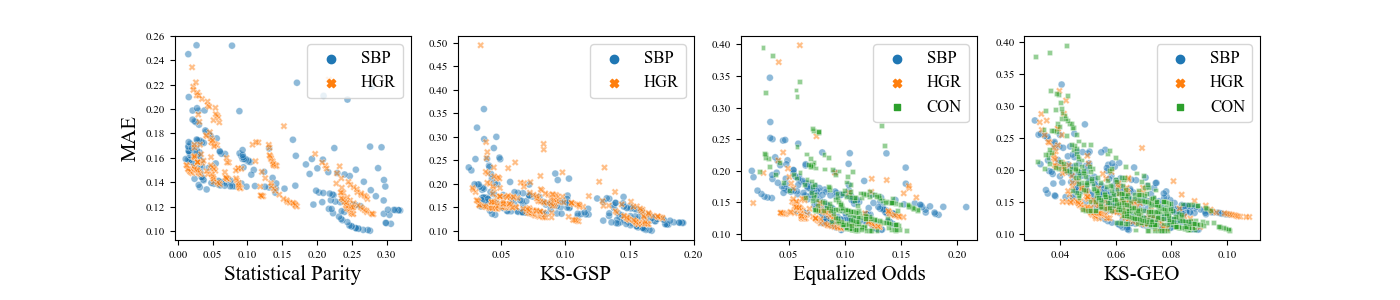}
    \caption{(Scenario III) Pareto frontiers: the points are pairs of the fairness metrics and MAE. Note the bottom-left tendency implies a better trade-off. }
    \label{sim:s3speo}
\end{figure}

In Scenario III, NEU cannot be implemented because NEU requires partitioning a data set with respect to the outcome variable, which is practically impossible for the continuous outcome. Figure~\ref{sim:s3speo} and Table~\ref{tab:sce3} show that SBP and HGR are comparable in Scenario III.  

\begin{table}[hbt!]
\caption{(Scenario III) Averages of the 5 smallest of SP/KS-GSP/EO/KS-GEOs whose MAEs are less than the thresholds. Those scores are selected in the Pareto solutions appearing in Figure~\ref{sim:s3speo}. Standard deviations are in the parentheses next to the averages.}
\label{tab:sce3}
\centering
\begin{tabular}{c|cc|cc}
\toprule
& \multicolumn{4}{c}{\textbf{Community and Crime} (MAE $\leq$ 0.12)}  \\
\hline
Model & SP   ($\downarrow$)    & KS-GSP  ($\downarrow$)   & EO   ($\downarrow$)        & KS-GEO     ($\downarrow$)       \\
\hline
\hline
SBP  & 0.225 (0.007)  & 0.141 (0.004) &  0.083 (0.003)      &  0.060 (0.002)  \\
CON  & -              & -             &  0.094 (0.004)      &  0.062 (0.002)  \\
HGR  & 0.247 (0.002)  & 0.155 (0.001) &  0.082 (0.004)      &  0.069 (0.002)  \\
\bottomrule   
\end{tabular}
\end{table}

\subsection{Robust estimation against the poor estimate of $\beta(a,y)$}
\label{sec:robust_beta}

\begin{figure}[h]
    \centering
    \includegraphics[trim={2.5cm 0 0 0},clip,scale=0.40]{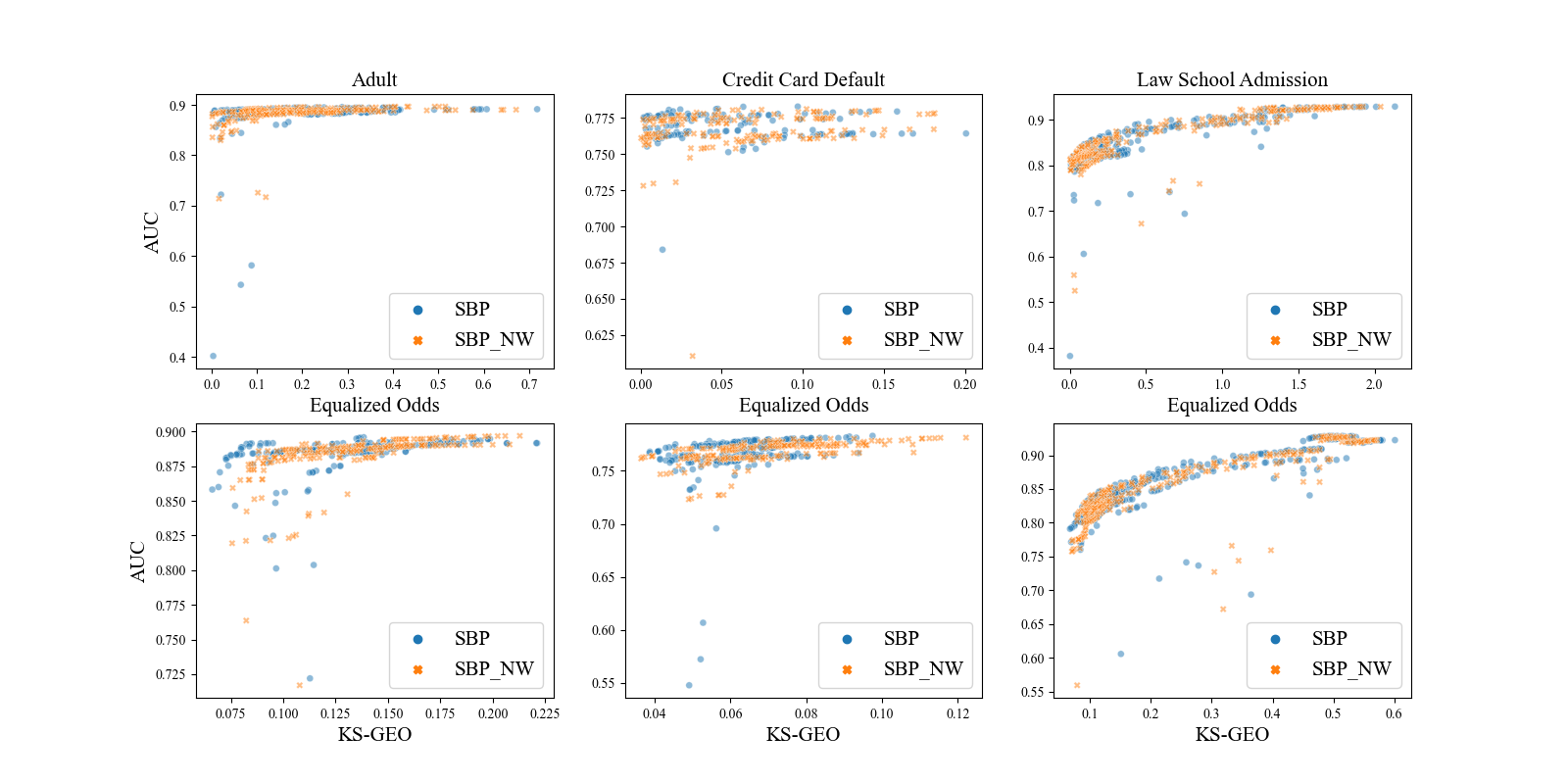}
    \caption{(Scenario I, the impact of $\hat{\beta}$) Pareto frontiers: the points are pairs of EO and AUC or KS-GEO and AUC.}
    \label{sim:s1eonw}
\end{figure}

\begin{figure}[ht]
    \centering
    \includegraphics[trim={2.5cm 0 0 0},clip,scale=0.45]{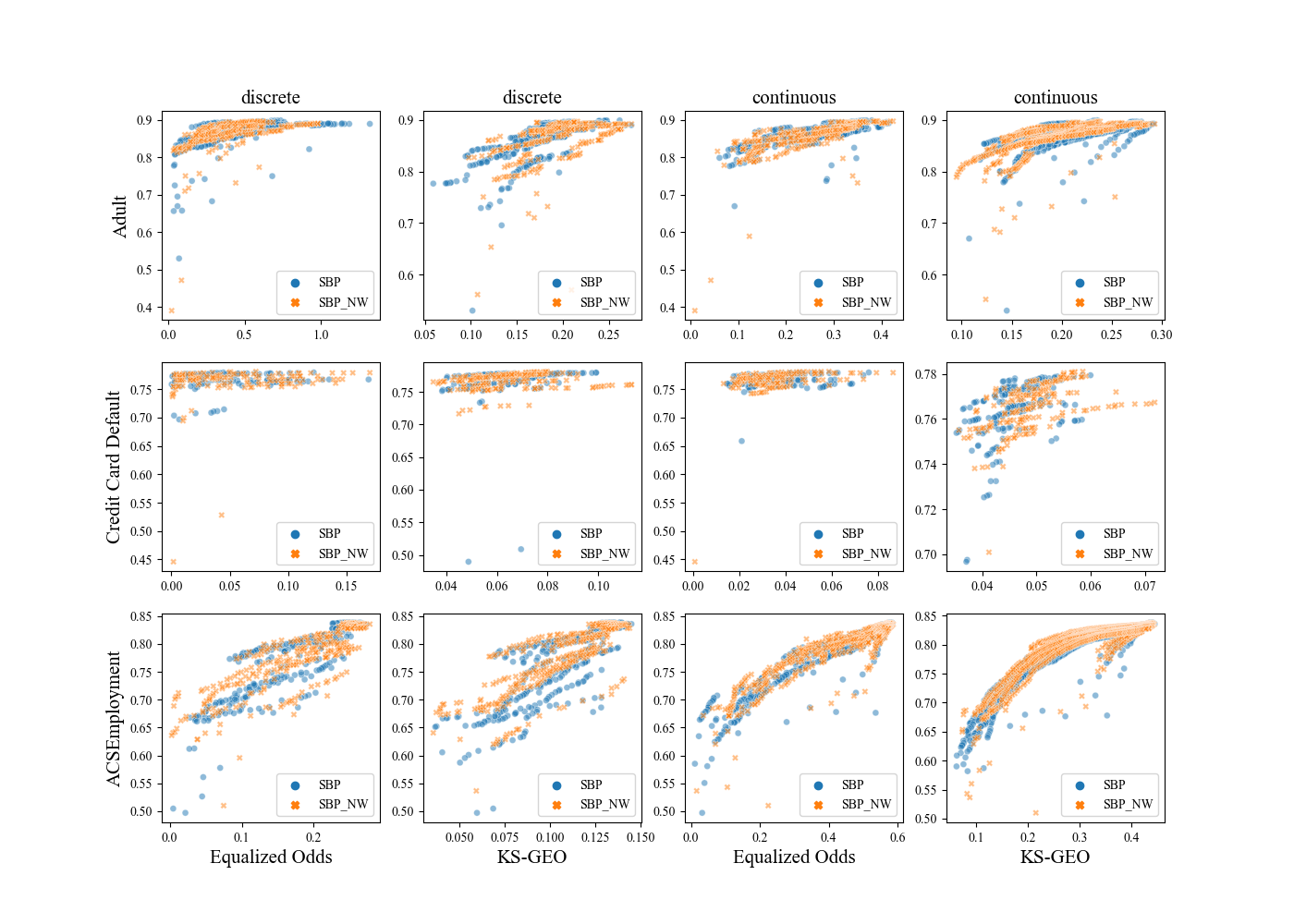}
    \caption{(Scenario II, the impact of $\hat{\beta}$) Pareto frontiers: the points are pairs of EO and AUC or KS-GEO and AUC. }
    \label{sim:s2speonw}
\end{figure}

We also investigate the impact of $\hat{\beta}$ discussed in Section~\ref{sec:sep} based on II. Interestingly, we observe that $\hat{\beta}$ hardly affects the performance of SBP even though the density-ratio estimator is required to guarantee GEO theoretically. Figure \ref{sim:s2speonw} compares the Pareto frontiers when using $\hat{\beta}(a,y)=\hat D_{\beta}(a,y)/(1-\hat D_{\beta}(a,y))$ (SBP) or using $\hat{\beta}(a,y)=1$ for all $a,y$  (SBP\_NW). SBP\_NW is set to having the same simulation configuration as SBP. The figure describes that SBP and SBP\_NW are almost the same in both scenarios. The same tendency is also found in Scenario I (Figure~\ref{sim:s1eonw}). 

\subsection{Fair Representation}
\label{sec:fair_rep}
We carry out additional simulation studies to verify that the neural penalty can be used to produce fair representation in Scenario II for EO. Following the same notation in Section~\ref{sec:sep}, we set $E$ as [Dense(64)-BN-ReLU] and $h_E$ as [Dense(64)-BN-ReLU]*2-[Dense(1)-Sigmoid], so that $h=h_E \circ E$. This is the same architecture as NEU. For HGR, we calculate $\text{HGR}_{\text{soft}}(E(X),A)$ for SP and ${\text{HGR}_{\text{soft}}}(E(X),A\otimes Y)- {\text{HGR}_{\text{soft}}}(E(X), Y)$ for EO. When training SBP and HGR, the number of training iterations for the maximization part is set to 5, i.e., $T'=5$. 

\begin{figure}[h]
    \centering
    \includegraphics[trim={2.5cm 0 0 0},clip,scale=0.45]{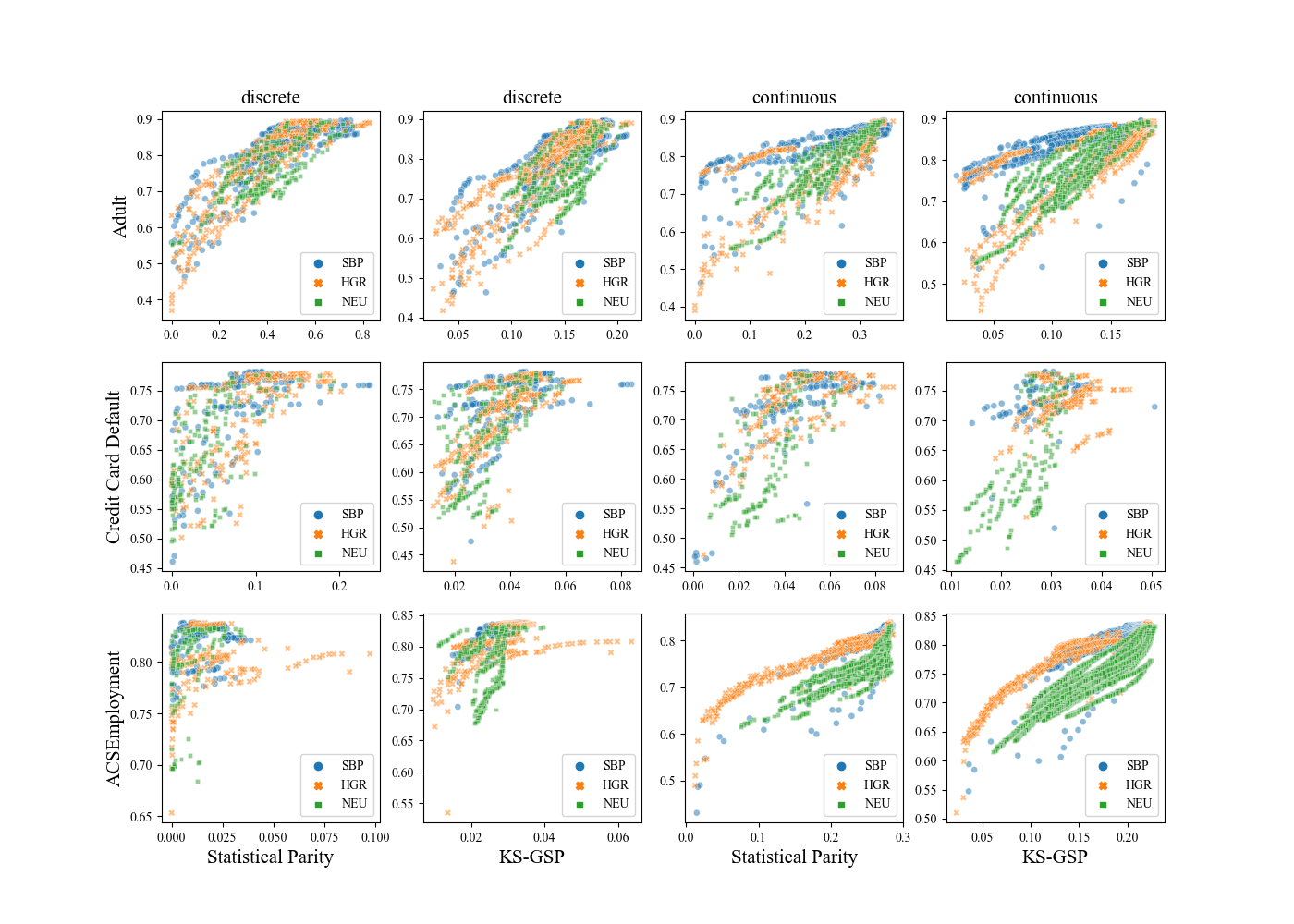}
    \caption{(Scenario II, fair representation) Pareto frontiers: the points are pairs of SP and AUC or KS-GSP and AUC.}
    \label{sim:s2spl}
\end{figure}

\begin{figure}[h]
    \centering
    \includegraphics[trim={2.5cm 0 0 0},clip,scale=0.45]{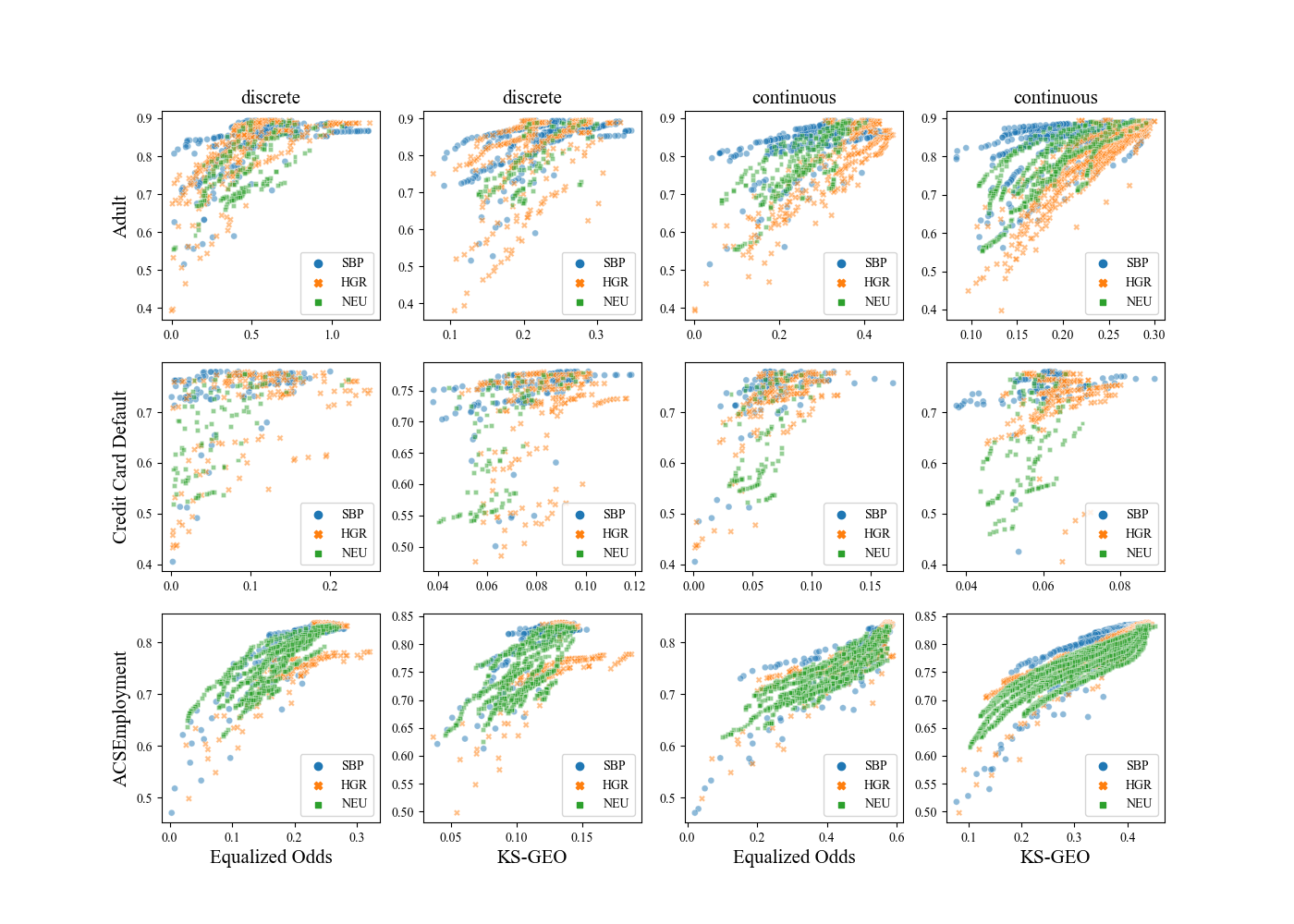}
    \caption{(Scenario II, fair representation) Pareto frontiers: the points are pairs of EO and AUC or KS-GEO and AUC.}
    \label{sim:s2speol}
\end{figure}

Figure~\ref{sim:s2speol} highlights that SBP succeeds in generating fair representation as achieving comparable or even better performance than the competing methods. The collected Pareto frontiers of the models on the benchmark data sets illustrate that SBP tends to defeat others slightly. We see that the SBP can represent the fair representation for SP as well (Figure~\ref{sim:s2spl}). 

\end{document}